\documentclass[preprint,12pt,authoryear]{elsarticle}

\usepackage{amssymb}
\usepackage{amsmath}
\usepackage{natbib}
\usepackage{bm}
\usepackage{booktabs}
\usepackage{multirow}
\usepackage{array}
\usepackage{threeparttable}

\begin{document}

\begin{frontmatter}



\title{Pedestrian crossing decisions can be explained by bounded optimal decision-making under noisy visual perception}

\author[inst1]{Yueyang Wang}
\author[inst1]{Aravinda Ramakrishnan Srinivasan}
\author[inst2]{Jussi P.P. Jokinen}
\author[inst3]{Antti Oulasvirta}
\author[inst1]{Gustav Markkula}

\affiliation[inst1]{organization={Institute for Transport Studies},
            addressline={University of Leeds}, 
            city={Leeds},
            postcode={LS2 9JT}, 
            state={Leeds},
            country={UK}}

\affiliation[inst2]{organization={Faculty of Information Technology},
            addressline={University of Jyväskylä}, 
            city={Jyväskylä},
            postcode={40100},
            country={Finland}}

\affiliation[inst3]{organization={Department of Information and Communications Engineering},
            addressline={Aalto University}, 
            city={Espoo},
            postcode={00076}, 
            country={Finland}}
\vspace{-2cm}

\begin{abstract}

This paper presents a model of pedestrian crossing decisions, based on the theory of computational rationality. It is assumed that crossing decisions are boundedly optimal, with bounds on optimality arising from human cognitive limitations. While previous models of pedestrian behaviour have been either 'black-box' machine learning models or mechanistic models with explicit assumptions about cognitive factors, we combine both approaches. Specifically, we model mechanistically noisy human visual perception and assumed rewards in crossing, but we use reinforcement learning to learn bounded optimal behaviour policy. The model reproduces a larger number of known empirical phenomena than previous models, in particular: (1) the effect of the time to arrival of an approaching vehicle on whether the pedestrian accepts the gap, the effect of the vehicle's speed on both (2) gap acceptance and (3) pedestrian timing of crossing in front of yielding vehicles, and (4) the effect on this crossing timing of the stopping distance of the yielding vehicle. Notably, our findings suggest that behaviours previously framed as ’biases’ in decision-making, such as speed-dependent gap acceptance, might instead be a product of rational adaptation to the constraints of visual perception. Our approach also permits fitting the parameters of cognitive constraints and rewards per individual, to better account for individual differences. To conclude, by leveraging both RL and mechanistic modelling, our model offers novel insights about pedestrian behaviour, and may provide a useful foundation for more accurate and scalable pedestrian models. 
\end{abstract}


\end{frontmatter}

\newpage
\section{Introduction}
\label{sec:Introduction}

\subsection{Background}
\label{subsec:Background}


Pedestrian crossing behaviour is a key factor in the urban transport system,  impacting traffic efficiency and safety \citep{leu2012modeling}. Developing accurate models for pedestrian crossing decisions is therefore essential for effective urban planning \citep{gonzalez2021agent,pelorosso2020modeling}. In particular, with the rise of autonomous vehicles (AVs), understanding pedestrian behaviour is important for the safe and harmonious integration of AVs into the existing transport ecosystem \citep{gonzalez2021agent,pelorosso2020modeling,duric2002integrating,banovic2019computational,al2001framework,li2016human}. The unpredictable nature of pedestrian movements, influenced by a complex interplay of cognitive processes and environmental cues, presents a significant challenge in this regard \citep{camara2020pedestrian,crosato2023social}.

In pedestrian behaviour modelling, computational models are predominantly based on either mechanistic modelling or on data-driven machine learning (ML) approaches. Mechanistic models, which are grounded in assumptions about underlying cognitive mechanisms causing the observed behaviours, aim to accurately represent the underlying processes of pedestrian behaviour \citep{fajen2003behavioral,turnwald2016understanding,yang2020social,pekkanen2021variable}. However, formulating these mechanisms in a way that generalises across a wide variety of different situations and scenarios is highly non-trivial. For this reason, mechanistic models typically suffer from scalability or generalisability issues due to the complexities in real-world traffic conditions. On the other hand, data-driven ML models, which predict pedestrian movement by learning from large datasets of pedestrian trajectories, show promise in handling
complex real-world situations \citep{dai2019modeling,abughalieh2020predicting,quan2021holistic,yin2021multimodal,yuan2021agentformer,zhang2022st,li2022graph}. However, they come with inherent challenges related to interpretability, data dependence, and robustness in diverse conditions \citep{althoff2018automatic,klischat2020scenario}. These issues will be discussed in more detail in Section \ref{subsec:Related work}.

Ideally, one would want a modelling approach which can combine the strengths of both mechanistic and ML models. One possible candidate for such an approach is available from \emph{computational rationality} \citep{lewis2014computational,gershman2015computational}, sometimes also referred to as resource rationality \citep{lieder2020resource}. Both computational rationality and resource rationality are theories about human cognition and behaviour, positing that humans behave \emph{rationally}, i.e., \emph{optimally} with respect to some utility or cost function, but that this optimality is \emph{bounded} by the constraints imposed by the human cognition and body. For example, pedestrians in road traffic have perceptual, cognitive, and motor limits which constrain their behaviour. Rooted in the principle of expected utility maximization, introduced by 
\citet{von1947theory}, and expanded by Herbert Simon's concept of bounded rationality \citep{simon1955behavioral}, computational rationality applies these decision-making principles while incorporating the realistic limitations faced by individuals. In recent years, the development of models based on these theories has become increasingly attainable, due to advances in the field of deep reinforcement learning (RL), enabling the learning of optimal decisions through environmental interactions \citep{silver2016mastering,silver2017mastering,sutton2018reinforcement}. Modern deep RL allows us to find boundedly optimal behaviour policy across highly varied scenarios and learn interactive behaviours which--to the extent that the theory of computational rationality is correct--mirror human-like decision processes and behavioural adaptability \citep{jokinen2020adaptive,jokinen2021multitasking,chen2021apparently,chen2021adaptive}. Contrary to traditional mechanistic models, which often struggle with flexibility in adapting to diverse scenarios and integration of various principles, and ML models, which can lack interpretability and adaptability, the computational rationality framework, when combined with RL, has the potential to overcome these limitations \citep{oulasvirta2022computational}.

There have been some initial studies on the use of computational rationality in driver behaviour modelling~\citep{jokinen2021multitasking,jokinen2022bayesian}, but so far none in the context of vulnerable road users. Here, we provide a first demonstration of the value of this modelling framework in the context of pedestrian behaviour. We test our model on data from a controlled virtual reality experiment on pedestrian road-crossing. The main human limitation that we are considering in this work is imperfect human visual perception, characterised by the inherent noise in the visual system's processing of dynamic stimuli. This limitation often leads to mistakes in how people estimate the distance and speed of oncoming traffic, which we model mechanistically based on work in cognitive neuroscience. We also propose a simple reward function for pedestrian crossing, which includes a concept of visual looming discomfort. Moreover, we show how individual differences in road user behaviour can be efficiently modelled using the computational rationality approach, by conditioning the RL on the human constraint parameters. This method, which has not been previously used in road user behaviour modelling, includes parameters which can vary between humans as inputs to the RL, to learn optimal policy across variations in these parameters.

This paper is organised as follows: The rest of this section describes empirical studies of the behavioural phenomena we wish to capture, as well as previous research on pedestrian behaviour modelling. Section 2 describes the dataset we used and the modelling approach. The results of our proposed model are presented in Section 3, followed by Section 4 with a discussion of results and future research plans. Finally, Section 5 provides a conclusion.

\subsection{Related work}
\label{subsec:Related work}

Numerous empirical studies have investigated the key factors influencing pedestrian behaviour \citep{oxley2005crossing,jain2014pedestrian,sun2015estimation,asaithambi2016pedestrian,gorrini2018observation,tian2022explaining}. We focus on four main empirical phenomena, some but not all of which have been captured by existing models: 

\emph{(1) TTA-dependent gap acceptance:} A key behaviour of interest is \emph{gap acceptance}, where pedestrians decide to cross based on the time or spatial distance between them and an approaching vehicle. One crucial factor affecting this decision is the \emph{time to arrival (TTA)}. Researchers have shown that pedestrians are more inclined to accept gaps for crossing when the TTA is higher \citep{oxley2005crossing,lobjois2007age,petzoldt2014relationship}. 

\emph{(2) Speed-dependent gap acceptance:}
Additionally, the dynamics of approaching vehicles play an important role \citep{schneemann2016analyzing}. Pedestrians have been found to be more likely to cross in a given TTA when oncoming vehicles travel at higher speeds, in other words with a larger approach distance \citep{lobjois2007age,tian2022explaining}. This phenomenon has been described as a bias in the human estimation of TTA \citep{petzoldt2014relationship,sun2015estimation}. \cite{tian2022explaining} explained this speed-dependent crossing behaviour by visual looming, the perceived growth of an object's size as it approaches \citep{delucia2008critical}. Specifically, visual looming increases slowly at long distances, indicating that higher speed vehicles might produce smaller collision threats to pedestrians for a chosen TTA, thus influencing pedestrians to feel it is safer to cross.

\emph{(3) Speed-dependent yielding acceptance:} In scenarios where vehicles yield, the relationship between speed and pedestrian crossing behaviour tends to reverse. Pedestrians have been found to interpret lower vehicle speeds as an indication of yielding, especially when the vehicle is at a closer distance, thus increasing the probability of crossing \citep{tian2023deceleration}. On the other hand, early crossing decisions (while the vehicle is still some distance away) are similar to the speed-dependent gap acceptance regardless of whether the car is yielding or not.

\emph{(4) stopping distance-dependent yielding acceptance:} Driver behaviour, particularly the use of exaggerated deceleration or 'short-stopping' serves as a cue for pedestrians to cross \citep{domeyer2019proxemics}. This action enhances the pedestrian’s perception of the driver's intent, increasing their confidence in crossing safely \citep{domeyer2019proxemics}. \cite{Risto2017Human-Vehicle} argued that the stopping distance of a vehicle correlates with pedestrian willingness to cross in the yielding scenario. Notably, \cite{tian2023deceleration} found that the car's braking behaviour mainly influences late crossing decisions. Specifically, pedestrians tend to cross more readily when a vehicle stops at a greater distance from the crosswalk, seemingly interpreting this as a strong indication of the driver’s intent to yield. 


\begin{table}[!b]
\centering
\caption{Key phenomena in pedestrian behaviour and corresponding literature}
\label{tab:key_phenomena}
\begin{tabular}{>{\centering\arraybackslash}m{6cm} >{\centering\arraybackslash}m{3.5cm} >{\centering\arraybackslash}m{3cm}}
\toprule
\textbf{Phenomenon}& \textbf{\citet{petzoldt2014relationship,tian2022explaining}}  & \textbf{\citet{pekkanen2021variable}} \\
\midrule
(1) TTA-dependent gap acceptance & \checkmark & \checkmark \\
(2) Speed-dependent gap acceptance & \checkmark & \checkmark \\
(3) Speed-dependent yielding acceptance & & \\
(4) Stopping distance-dependent yielding acceptance & & \checkmark \\
\bottomrule
\end{tabular}
\end{table}

In our study, we aim to model pedestrian behaviour and reproduce these four above-mentioned phenomena, which are summarised in Table \ref{tab:key_phenomena}. These phenomena provide valuable insights and highlight some of the nuances of human decision-making in traffic interactions that computational models should capture in order to be useful for practical applications such as simulation environments and AV algorithms. As mentioned, there are many models which describe pedestrian behaviour in terms of hypothesised underlying mechanisms \citep{fajen2003behavioral,giles2019zebra,pekkanen2021variable,tian2022explaining,tian2023deceleration}. For example, \citet{petzoldt2014relationship} developed a logit-based gap acceptance model capturing speed-dependent gap acceptance, described as a bias due to a visual heuristic used by pedestrians. However, it remains unclear whether this heuristic is truly biased in the sense that it is suboptimal, or whether it might be an optimal adaptation to human limitations. Later, \cite{tian2022explaining} proposed a gap acceptance model utilising the binary choice logit approach. Different from previous models that relied on traffic gap cues, their model incorporated visual looming cues, which resulted in an improved fit to observed data. Another modelling approach is based on the concept of evidence accumulation, describing decision-making as a noisy accumulation of evidence from a stimulus \citep{ratcliff2016diffusion}, which has also been applied to pedestrian crossing behaviour modelling \citep{pekkanen2021variable}. However, due to the complexity of this model, the authors fitted it with a single parameterisation across all participants, and with limited quantitative goodness of fit to the experimental data. The complexity of this model also makes it difficult to extend it to more sophisticated scenarios. To extend the scope of modelled scenarios, \citet{markkula2023explaining} integrated a large number of existing psychological and cognitive theories, such as theories of sensory noise, Bayesian perception, evidence accumulation decision-making, and long-term valuation of action affordances. Although this model was capable of reproducing several empirically observed phenomena in human road user interaction, the authors highlighted the limitations of mechanistic modelling and underscored the need for cognitively and behaviourally informed ML. Overall, given the high complexity of human behaviour, relying on a single cognitive, mechanistic model seems insufficient to comprehensively describe pedestrian behaviour. The existing models' capture of the targeted phenomena is listed in \tablename~\ref{tab:key_phenomena}. However, it is worth noting that the aim of our work is not only to capture these phenomena but also to propose a general framework for modelling pedestrian behaviour.

With advances in computing power and machine-learning methods, a growing number of researchers have adopted data-driven ML algorithms for modelling pedestrian behaviour. For instance, many models based on Convolutional neural networks have been developed for pedestrian trajectory prediction due to their ability to process spatial inputs, such as images and video frames. \citep{yi2016pedestrian,abughalieh2020predicting,kumamoto2017cnn,doellinger2018predicting}. Moreover, recurrent neural networks have garnered attention given their efficacy in sequence prediction tasks, making them apt for pedestrian trajectory modelling \citep{alahi2016social,dai2019modeling,quan2021holistic}. Recently, the Transformer model, known for its ability to process sequences in parallel and capture long-range interactions, has also gained popularity in pedestrian behaviour modelling \citep{lorenzo2021capformer,yin2021multimodal,yuan2021agentformer}. While the accuracy of these ML models can be impressive, they share several limitations. First, they often act as 'black boxes', meaning they may have the potential to predict behaviours but fail to reveal the reasons or mechanisms behind those behaviours. This intensive data-driven orientation can be a challenge, especially when aiming for interpretability in AV systems where understanding pedestrian intent is crucial \citep{srinivasan2023beyond}. Second, the lack of theoretical grounding of the behaviour and minimising trajectory prediction error can sometimes lead to overfitting, where the models might perform well on the training data but fail to generalise well to unseen data, potentially resulting in less reliable predictions in varied real-world scenarios \citep{xu2020explainable}.  Third, their heavy reliance on large training datasets poses another challenge: collecting such extensive data under all possible road conditions is almost an impossible task, and critical scenarios are missing in most datasets \citep{diaz2022ithaca365}.

As mentioned in Section \ref{subsec:Background}, computational rationality provides a potential approach to combining the best of mechanistic and ML models but has only seen limited use in road user behaviour modelling. \citet{jokinen2021multitasking} proposed a hierarchical RL model of multitasking behaviour in driving and fitted it to observed data \citep{jokinen2022bayesian}. However, their focus was on in-vehicle multitasking rather than pedestrian-vehicle interactions. Here, we aim to quantitatively model the crossing decisions of human pedestrians when faced with constant-speed and yielding vehicles. There are three research objectives in this study:

(1) To develop a bounded optimal model, integrating human constraints, with a focus on noisy perception and looming aversion, that is able to capture the four targeted empirical phenomena in human pedestrian crossing behaviours, in scenarios where vehicles may or may not yield to the pedestrian.

(2) To fit the bounded optimal model quantitatively to human crossing decision data on a per-individual level.

(3) To assess how the human constraints and preferences integrated into the model influence the crossing decisions.

An early version of this work, partially addressing the first objective above (two of the four empirical phenomena), but not the second or third objectives, has been presented as a conference paper \citep{wang2023modeling}. 

\begin{figure}[!b]
      \vspace{-0.2cm}
      \centering
      \includegraphics[scale=0.5]{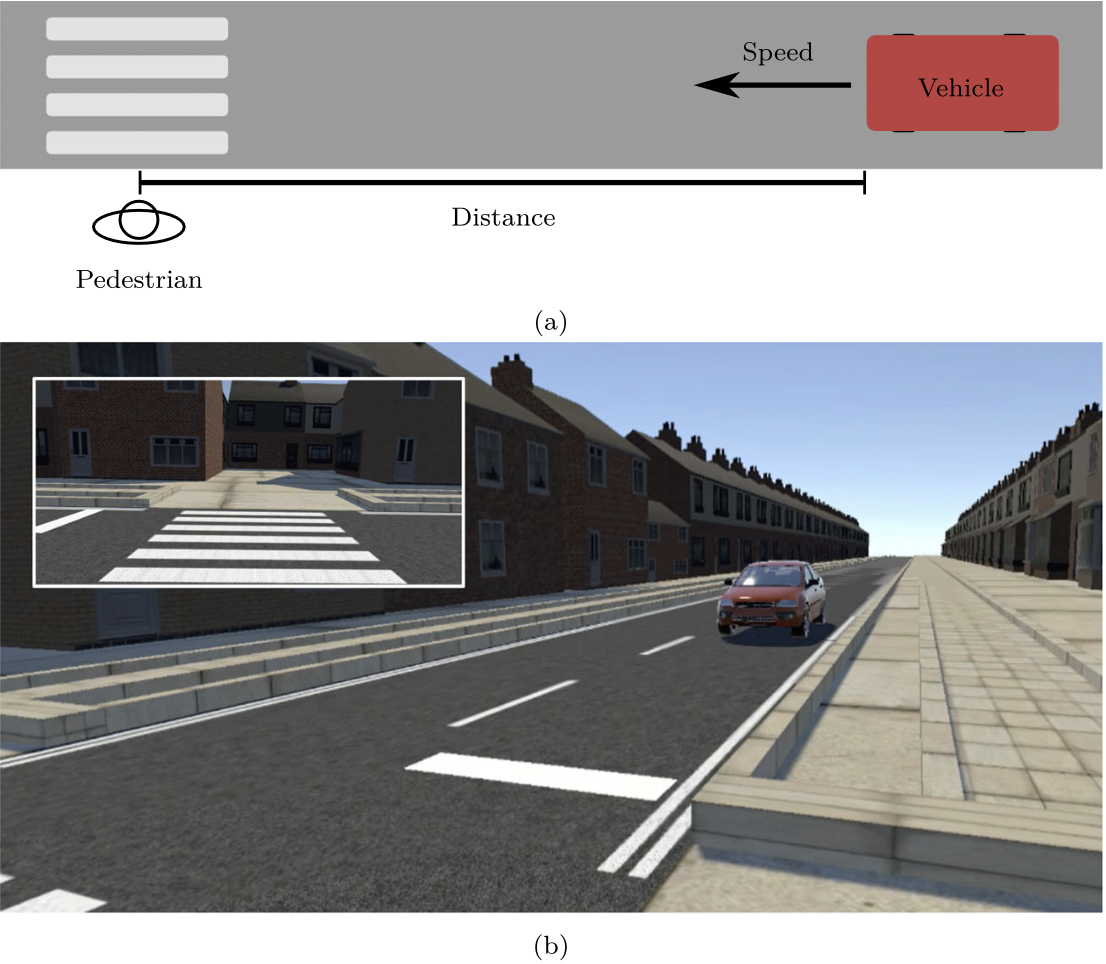}
      \caption{(a) Birds-eye view of the experiment. (b) A sample view of the virtual scene, as shown by the head-mounted display, at the beginning of each trial (inset) and as participants turned their heads to look for oncoming traffic.
      (Source: Pekkanen et al., 2021. Variable-drift diffusion models of pedestrian road-crossing decisions. Computational Brain $\&$ Behavior, 1–21. This image is available under a Creative Commons Attribution 4.0 International License.)}
      \label{fig:Birds_eye_view}
      \vspace{-0.2cm}
\end{figure}

\section{Methods}
\label{sec:Methods}

\subsection{Dataset}
\label{subsec:Dataset}
This study utilised a dataset sourced from a previous experiment reported by \citet{giles2019zebra}. A visualisation of the experimental setup is provided in \figurename~\ref{fig:Birds_eye_view}, which offers a bird's-eye perspective. A total of 20 participants were recruited for this experiment. As part of the setup, participants wore an HTC Vive Virtual Reality (VR) headset, immersing them in a virtual crossing task. The rendered VR space included a straight two-lane road, spanning a width of $5.85$~m, and a zebra crossing at the participant's initial location. 

\begin{table}[!b]
    \vspace{-0.4cm}
    \caption{Detailed description of vehicle approach scenarios in the experiment. The table lists each scenario type along with key parameters: Initial Vehicle Speed ($v_0$ in $m/s$), Initial Distance ($d_0$ in $m$) from the pedestrian, Initial TTA ($\tau_0$ in $s$), and Stopping Distance ($d_\mathrm{stop}$ in $m$) for Yielding scenarios. 'N/A' indicates not applicable for constant speed scenarios.}
    \centering
    \begin{tabular}{ccccc}
        \toprule
        \textbf{Scenario type} & \bm{$v_0~(m/s)$} & \bm{$d_0~(m)$} & \bm{$\tau_0~(s)$} & \bm{$d_\mathrm{stop}~(m)$} \\
        \midrule
        Constant speed  & $6.94$  & $15.90$ & $2.29$ & $N/A$ \\
        $ $  & $13.89$ & $31.81$ & $2.29$ & $N/A$ \\
        $ $  & $6.94$  & $31.81$ & $4.58$ & $N/A$ \\
        $ $  & $13.89$ & $63.61$ & $4.58$ & $N/A$ \\
        $ $  & $6.94$  & $47.71$ & $6.87$ & $N/A$ \\
        $ $  & $13.89$ & $95.42$ & $6.87$ & $N/A$ \\
        Yielding  & $6.94$  & $15.90$ & $2.29$ & $4$ \\
        $ $  & $13.89$ & $31.81$ & $2.29$ & $4$ \\
        $ $  & $13.89$ & $31.81$ & $2.29$ & $8$ \\
        $ $  & $6.94$  & $31.81$ & $4.58$ & $4$ \\
        $ $  & $13.89$ & $63.61$ & $4.58$ & $4$ \\
        $ $  & $13.89$ & $63.61$ & $4.58$ & $8$ \\
        $ $  & $6.94$  & $47.71$ & $6.87$ & $4$ \\
        $ $  & $13.89$  & $95.42$ & $6.87$ & $4$ \\
        \bottomrule
    \end{tabular}       
    \vspace{-0.4cm}
    \label{tab:scenarios}
\end{table}

At the start of each trial, participants were positioned before the zebra crossing facing straight across it. They had been instructed that when they felt ready to begin the trial, they should look to the right for any oncoming traffic. This head movement triggered the start of the scenario, with an oncoming vehicle initialised at an initial distance $d_0$ and speed $v_0$. The experiment included a mix of scenarios, namely constant-speed and yielding scenarios. The detail of these scenarios is shown in \tablename~\ref{tab:scenarios} with the initial TTA $\tau_0 = d_0/v_0$ also listed. In six constant-speed scenarios, the vehicle appeared at a distance $d_0$ (all distances measured longitudinally along the road from the participant’s location to the front of the car) and maintained a constant-speed $v_0$ while approaching and passing the zebra crossing. In eight yielding scenarios, the vehicle appeared at initial distance and speed $v_0$ and $d_0$, and immediately decelerated at a constant rate to stop at a distance $d_\mathrm{stop}$ from the participant. Participants were instructed to press the button on the HTC Vive’s controller when they felt safe to cross. It should be noted that the vehicle’s behaviour was not influenced by the pedestrian's decision. This setup required participants to independently assess crossing safety. Once this button was pressed, the system recorded the \emph{Crossing Initiation Time (CIT)}, defined as the interval between the start of each trial and the moment when the button press occurred. At this point, the position of the participant's point of view in the virtual environment (i.e., the VR 'camera') moved across the zebra crossing at the speed of $1.31$~m/s. It is worth noting that the decision to adopt this button-press mechanism, as opposed to actual physical road crossing, was to circumvent between-participant variability arising from varying motor abilities influencing the crossing decision. Each participant faced 6 unique constant-speed trials and 8 unique yielding trials in a randomised order, aggregating to a total of 280 trials that were used for the model's validation.

The low number of repetitions per participant was adopted by \citet{giles2019zebra} to limit behavioural adaptation effects. The resulting dataset is relatively small in size, but since all of the main phenomena we are targeting have been reported also in other experiments, the sample size is not a main concern here. 
We adopted this dataset here because of the button-press paradigm, which provides crossing onset distributions with minimal impact from motor variability, aligning with our present modelling emphasis on perception rather than motor control.

\subsection{Model variants}
\label{subsec:Main Model variants}

In this section, we explain our two main mechanistic assumptions, rooted in cognitive science and neuroscience, and introduce different model variants derived from these assumptions:

\begin{itemize}
    \item \textbf{Assumption of noisy perception:} We assume that the agent's perception of the environment is inherently noisy and imperfect. This noisy observation obtained by the agent is according to the principle of the human visual system, i.e., the sensory input received by our human visual system is noisy \citep{faisal2008noise}.

    \item \textbf{Assumption of looming aversion:} Our second assumption involves the natural aversion to looming objects \citep{delucia2008critical,tian2022explaining}. Looming aversion refers to the instinctive tendency to avoid objects that appear to be rapidly growing in size, as this is often a cue for an impending collision \citep{delucia2008critical,tian2022explaining}.
\end{itemize}

These two assumptions guide the development of our model variants. Each variant incorporates the assumptions in different ways to explore how they individually and collectively influence pedestrian behaviour. The following subsection outlines the four main model variants derived from these assumptions, as visualised in \figurename~\ref{fig:Model}:

\begin{figure}[!t]
      \vspace{-0.4cm}
      \centering
      \includegraphics[scale=0.5]{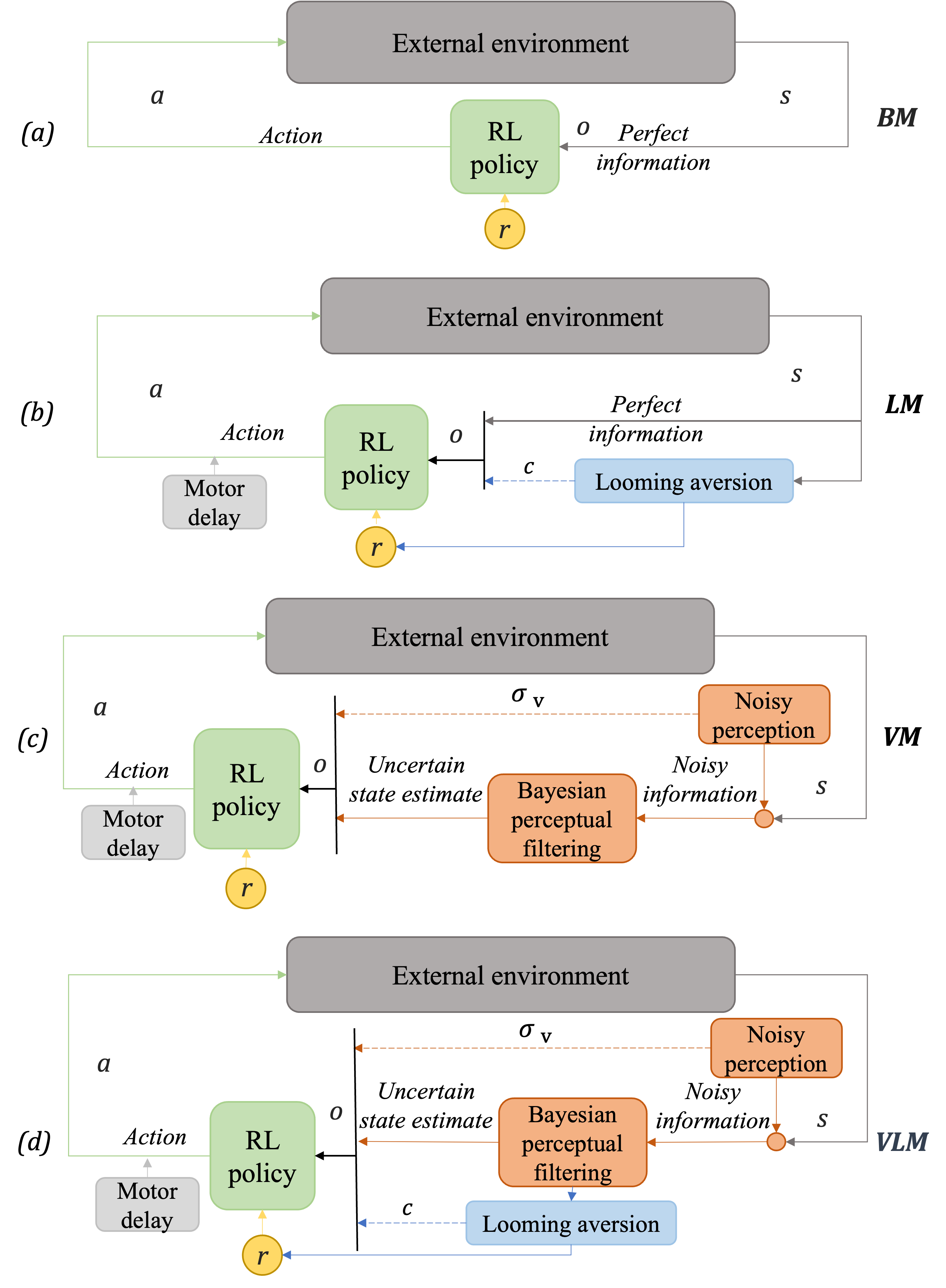}
      \caption{Comparison of models. (a) BM: Baseline Model. (b) LM: Looming aversion only. (c) VM: Visual limitation only. (d) VLM: Looming aversion and visual limitations. The abbreviations were introduced in Section~\ref{subsec:Main Model variants}. $\sigma_\mathrm{v}$ and $c$ represent the sensory noise and looming aversion weight respectively. $s$, $o$, $a$, and $r$ represent the state, observation, action and reward respectively.}
      \label{fig:Model}
      \vspace{-0.4cm}
\end{figure}
(1) \textbf{Baseline Model (BM):} The BM serves as our control variant, assuming an ideal observer with neither visual limitations nor looming aversion. This model establishes a baseline against which to measure the impact of our two key assumptions.

(2) \textbf{Looming Model (LM):} The LM considers the impact of looming aversion on crossing decisions. It allows us to understand how this aversion, independent of visual limitations, can affect pedestrian behaviour.

(3) \textbf{Visual limitation Model (VM):} Conversely, the VM examines the role of visual limitations alone. It helps us investigate how the noisy sensory information influences pedestrian crossing.

(4) \textbf{Visual limitation and Looming Model (VLM):} The VLM combines both visual limitations and looming aversion to provide insights into their joint effect. 

Each combination of model assumptions mentioned above defines an RL problem. In the next section, we describe this RL problem in detail, for the different model variants.

\subsection{Reinforcement learning problem}
\label{subsec:Reinforcement learning problem}

In our model based on computational rationality, we used RL to derive the near-optimal behaviour under constraints. This approach differs from traditional data-driven ML algorithms, which typically learn directly from large datasets without iterative interaction with the environment. RL offers a paradigm wherein an agent interacts with a dynamic environment, and the optimal policy will be derived through trial-and-error~\citep {kaelbling1996reinforcement}. The RL environment is often modelled as a Markov Decision Process (MDP). This mathematical framework is represented by a tuple $<S,A,T,R>$, comprising a set of states $S$, a set of actions $A$, a transition function $T$, and a reward function $R$. The agent chooses the action by following a policy $\pi$, which yields a probability $\pi(s, a) = p(a \mid s)$ of taking a particular action from the given state. The optimal policy~$\pi^*$, maximises the value function:
\begin{equation}
V^*(s) = \max_{a} \left[ R(s, a) + \gamma \sum_{s' \in S} T(s, a, s') V^*(s') \right]
\end{equation}

where~$\gamma$~is the discount factor, a value between 0 and 1, which is used to balance the importance of immediate rewards versus future ones.

In our study, we assumed that human makes decisions under uncertainties from the noisy perception system, which can be described by a Partially Observable Markov Decision Process (POMDP), an extension of the MDP formalism. In a POMDP, the agent does not have direct access to the true state $S$; rather, it receives observations that may only partially or noisily reflect the actual state. The POMDP is represented by a tuple $<S,A,T,R,O>$, where $S$ is a set of states, $A$ is a set of actions, $T$ is the transition function, $R$ is the reward function and $O$ is the set of possible observations received by the agent.

\paragraph{State space $S$}

At each time step $t$, the environment is in a state $s_t \in S$. This is the position and velocity of both the pedestrian (agent) and the vehicle. All model variants in our study share the same state space, encapsulating these five critical variables: the pedestrian's position \(x_\mathrm{p}, y_\mathrm{p}\), the vehicle's position \(x_{\text{veh}}, y_{\text{veh}}\), the vehicle's velocity \(v\), and time step $t$. The simulation state update time step is set to 0.1 s, a duration suitable for our dataset and model.
 
\paragraph{Action space $A$}
At every time step $t$, the agent executes an action $a_t$ from the set $A$. For the purposes of this study, and consistent with the experimental button press, the agent's choices are binary: either to 'Go' or to 'Not Go'. Upon selecting 'Go', the agent proceeds straight at the speed of $1.31$ m/s, as in the experiment. Recognising that human reaction times vary, we introduced a motor delay to simulate this aspect. This delay, implemented after the 'Go' decision, was sampled from a Gaussian distribution with a mean of  $0.6$ s and a standard deviation of $0.2$ s. After the decision was made, the simulation progressed until the agent either crossed the road safely or a collision happened.

\paragraph{Transition $T$}
The transition function defines how the current state $s_t$ changes to the next state $s_{t+1}$ based on the action $a_t$. In our model, when the 'Not Go' action is selected, the vehicle's movement follows kinematic equations with the given speed, and the agent's position remains unchanged. Conversely, when the 'Go' action is chosen, whether the collision happens is calculated, and the corresponding reward is given to the agent. Then, the simulation finishes. 

\paragraph{Reward $R$}
In the experiment where the dataset was collected, the participant’s task was to cross the road as soon as they felt safe to do so, either before or after the car had passed them~\citep{giles2019zebra}. We are therefore assuming that the participants in this experiment (presumably similar to pedestrians in real traffic) wanted to cross safely, but with minimal loss of time, and we designed the reward structure accordingly: The agent will be given a reward of $20$ when crossing the road without collision, and a reward of $-20$ if a collision happens. These reward parameters were shared across model variants and were fixed after some initial manual testing, to yield reasonable crossing behaviour of the BM agent, but before structured fitting of any other model parameters to the human data. A time penalty, a negative reward of $0.01 \times t$, will also be given to the agent when the episode terminates.

Regarding the visual looming, this phenomenon is represented mathematically as inverse $\tau$—the ratio of a vehicle’s optical expansion rate to its size on the observer's retina, which serves as an estimate of the inverse TTA \citep{delucia_2015,markkula2016farewell}. We incorporated this concept into the reward function of the \emph{LM} and \emph{VLM} models, as shown in panels (b) and (d) of \figurename~\ref{fig:Model}, to account for the looming aversion in pedestrian decision-making.

The reward function $r$ is defined as follows:

\begin{equation}
\label{eq:reward}
r = 
\begin{cases}
\max(-20, \min(+20, 20 - 0.01 \cdot {t} - c \cdot \frac{1}{\hat{\tau}} )), & \text{if arrival} \\
-20, & \text{if collision}
\end{cases}
\end{equation}
where \( c \) is the weight of the looming aversion. For model variants with non-noisy perception, \({\tau} = \frac{x_\mathrm{veh} - x_\mathrm{p}}{v}\), and for those with noisy perception, \(\hat{\tau} = \frac{\hat{x}_\mathrm{veh} - x_\mathrm{p}}{\hat{v}}\), where $\hat{x}_\mathrm{veh}$ and $\hat{v}$ are noisy estimates of vehicle position and speed (see further below). The reward \( r \) is bounded within the range \([-20, +20]\) to prevent extreme values from influencing the model training. As the focus of this study is on the potential effect of noisy perception and the looming aversion on the crossing decision, we have kept the reward function simple; future work can further refine it to better capture human preferences.

\paragraph{Observation space $O$}
The agent receives observation $o_t \in O $ at each time step. We tested different formulations of the observation space: In \emph{BM} - our simplest, baseline model as shown in panel (a) of ~\figurename~\ref{fig:Model} - the exact position and velocity of both the vehicle and the pedestrian are presented as inputs to the agent. Conversely, in the models with visual limitation assumption, \emph{VM} and \emph{VLM}, illustrated in panels (c) and (d) of \figurename~\ref{fig:Model}, the position and velocity details provided to the agent are subject to noise, emulating the inherent uncertainty and imperfection in human visual perception. Regarding the nature of this noise, our models are based on the assumption that visual noise is introduced at the level of the human retina, as angular noise~\citep{kwon2015unifying}. Building on this understanding of visual noise, we assume that the agent observes the position of the other agent along its line of travel by observing the angle below the horizon of the other agent~\citep{ooi2001distance,markkula2023explaining}, with a constant Gaussian angular noise of standard deviation, $\sigma_\mathrm{v}$, which could vary between pedestrians. In practice, this means that the pedestrian observes the vehicle's distance along the road with a distance-dependent noise of standard deviation \citep{markkula2023explaining}:
\begin{equation}
\label{eq:noisy_perception}
\sigma_x(t) = |d_\mathrm{l}(t)| \left( 1-\frac{h}{d(t)\cdot\tan(\arctan\frac{h}{d(t)}+\sigma_\mathrm{v})}\right),
\end{equation}
\noindent where $d_\mathrm{l}(t)$ is the longitudinal distance between the vehicle and the crossing point, $d(t)$ is the distance between the agent and the approaching vehicle, and $h$ is the eye height over the ground of the ego agent, which is set to $1.6$~m for all pedestrian agents for simplicity.

Additionally, there is evidence that the human perception system interprets its noisy input in a Bayes-optimal manner, and Bayesian methods have been successful in modelling perception and sensorimotor control~\citep{kwon2015unifying,knill2004bayesian,stocker2006noise}. Therefore, we used a Kalman filter as a model of the human visual perception to perceive the environment~\citep{kwon2015unifying,markkula2023explaining}. In the initialisation of the Kalman filter, we used distinct prior distributions for position and velocity. Each distribution is centred on the actual vehicle state, with its standard deviation set equal to that of the initial set of values. At each time step, the Kalman filter receives noisy positional data regarding the other agent (vehicle). The filter then produces estimates of the vehicle's position and velocity, along with their respective uncertainties. These filtered estimates represent the agent's belief state, reflecting the Bayes-optimal inference of the vehicle's state based on the noisy observations. These belief estimates are then fed into the RL policy as inputs, together with the precise self-position and velocity of the agent itself, and the visual noise parameter $\sigma_\mathrm{v}$. 

Furthermore, the \emph{LM} and \emph{VLM} models incorporate the concept of visual looming. This looming aversion, modelled through inverse $\tau$ as shown in Equation~\ref{eq:reward}, affects the agent's perception of potential collision risks. We gave the weight for the looming aversion, \( c \), as input to the RL policy.

\begin{table}[t]
\centering
\caption{Observation space variables for each model variant. '\checkmark' indicates the variable is directly observed. Here, \(x_\mathrm{p}, y_\mathrm{p}\) denote the pedestrian's position, \(x_{\text{veh}}, y_{\text{veh}}\) the vehicle's position, \(v\) the vehicle's velocity, \(\hat{x}_{\text{veh}}, \hat{v}\) the Kalman filter's estimates of the vehicle's position and velocity, \(P_\mathrm{p}\) and \(P_\mathrm{v}\) are the variances in the Kalman filter's estimate of position and speed of the approaching vehicle, \(\sigma_\mathrm{v}\) represents the standard deviation of sensory noise, \(c\) the looming aversion weight, and \(t\) the time step.}
\begin{tabular}{>{\centering\arraybackslash}p{1cm} >{\centering\arraybackslash}p{0.7cm}  >{\centering\arraybackslash}p{0.6cm} >{\centering\arraybackslash}p{1.6cm} >{\centering\arraybackslash}p{1.6cm} >{\centering\arraybackslash}p{1.6cm} >{\centering\arraybackslash}p{3.7cm} }
\hline
\textbf{Model} & \( x_\mathrm{p} \) & \( y_\mathrm{p} \) & \( x_{\text{veh}} \) & \( y_{\text{veh}} \) & \( v \) & Additional Variables \\
\hline
BM    & \checkmark & \checkmark & \checkmark & \checkmark & \checkmark & \( t \) \\
LM    & \checkmark & \checkmark & \checkmark & \checkmark & \checkmark & \( c, t \) \\
VM    & \checkmark & \checkmark & Est.(\(\hat{x}_{\text{veh}}\)) & Est.(\(\hat{y}_{\text{veh}}\)) & Est.(\(\hat{v}\)) & \( P_\mathrm{p}, P_\mathrm{v}, \sigma_\mathrm{v}, t \) \\
VLM   & \checkmark & \checkmark & Est.(\(\hat{x}_{\text{veh}}\)) & Est.(\(\hat{y}_{\text{veh}}\)) & Est.(\(\hat{v}\)) & \( P_\mathrm{p}, P_\mathrm{v}, \sigma_\mathrm{v}, c, t \) \\
\hline
\end{tabular}
\label{tab:model_observation_spaces}
\end{table}

To distinguish $\sigma_\mathrm{v}$ and $c$ from the parameters of the policy neural network (connection weights and biases), we will refer to these two parameters as non-policy parameters. By feeding the non-policy parameters as inputs to the RL policy, we are not implying that the human agent 'observes' its own parameter values. Instead, we are just conditioning the RL policy on these non-policy parameters, as a more convenient alternative to learning entirely different policies for different parameter combinations.

\subsection{Reinforcement learning algorithm}
\label{subsec:Reinforcement learning algorithm}

Deep Q-Networks (DQNs), a class of RL algorithms, use neural networks to optimise the state-action value function, commonly known as the Q-function. The Q-function, denoted as $Q(s, a)$, is a fundamental concept in RL that quantifies the expected utility of taking a certain action $a$ in a given state $s$, and is defined as:
\begin{equation}
Q(s,a) = r + \gamma \max_{a'} Q(s',a'),
\end{equation}
where $s$ represents the current state, $a$ the action taken in state $s$, and $r$ the immediate reward received after taking action $a$ in state $s$. The discount factor $\gamma$ reflects the agent's weighting of future rewards: a $\gamma$ value close to 1 indicates a preference for long-term gains, whereas a value near 0 suggests a focus on immediate rewards.

The term $\max_{a'} Q(s',a')$ in the equation represents the maximum Q-value achievable in the next state $s'$ across all possible actions $a'$. This component is crucial as it enables the agent to estimate future rewards and informs the selection of the optimal action in each state. By continuously updating this Q-function through iterative learning and backpropagation, the DQN effectively learns a policy that guides the agent's decisions to maximise cumulative rewards. For a more detailed exploration of DQNs, please refer to \cite{mnih2015human}.

The DQN algorithm is suitable for problems with a continuous state space and a discrete action space, such as in our case here. We utilised an enhanced version of DQN, Double DQN (DDQN). DDQN's structure decouples the neural network parameter update for action selection from evaluation, mitigating overestimation of action value~\citep{van2016deep}. Additionally, we integrated the duelling network approach, partitioning the Q-function into a value function $V(s)$, representing the overall value of being in a given state, and an action advantage function $A(s,a)$, indicating the relative importance of each action in that state. This duelling structure is particularly beneficial in scenarios where different actions yield similar values; it enhances performance by considering both the value of the current state and the advantage of each action~\citep{wang2016dueling}.


We represented the $Q(s,a)$ values using a fully connected feedforward neural network with two hidden layers, composed of $512$ and 256 nodes. It has been observed that larger networks tend to yield more stable policies and policies are more robust to noise \citep{xie2019iterative}. The learning rate and discount factor were set to $0.0001$ and $0.99$ respectively. To encourage exploration, we implemented an $\epsilon$ - greedy algorithm: at each time step $t$, the system either randomly selects an action with probability $\epsilon$ or chooses the action with the highest Q value with probability $1-\epsilon$ \citep{wunder2010classes}. We initialised $\epsilon$ at 1 and decreased it by by $5^{-5}$ in each learning step. The minimum value of $\epsilon$ was set to $0.001$. We trained model M, VM and LM over 25,000 episodes for convergence. Convergence here means that rewards stabilise and no longer increase over time. The choice of this criterion is rooted in the understanding that the model has likely learned an optimal or near-optimal policy once the rewards stop improving significantly. For the model VLM, we extended the training to 45,000 episodes due to its added complexities from both visual limitations and looming aversion assumptions. 

To avoid the agent learning a simplistic strategy of always crossing immediately, we also included scenarios with an initial TTA of 1 s, in which safe crossing was not feasible before the vehicle's arrival.

\subsection{Fitting of non-policy parameters}
\label{subsec:Fitting of non-policy parameters}
Our most complex model variant VLM has two free model parameters, $\sigma_\mathrm{v}$ and $c$. Testing different values for these non-policy parameters is essential because they alter the environment in the RL problem, thus influencing the agent's crossing decision.

Learning a separate RL policy for each possible combination of non-policy parameter values is computationally expensive \citep{Howes2023Towards,Li2023Modeling}. To address this challenge, recent studies have adopted a more efficient approach within the computational rationality framework. This method involves conditioning the RL policy on non-policy parameters by integrating them as additional inputs to the model \citep{keurulainen2023amortised}. In this case, we provided $\sigma_\mathrm{v}$ and $c$ as additional inputs to the RL policy during learning, as illustrated by the dashed line in~\figurename~\ref{fig:Model} (d). Utilising this approach, the model can learn boundedly optimal policy for arbitary values of the non-policy parameters.

We do not know the correct values for the noise magnitude parameter, $\sigma_\mathrm{v}$, and the weight for the looming aversion, $c$, during the training phase. Moreover, these values might differ for individual participants in the experiment. Therefore, we trained the RL policy for each model variant across a range of $\sigma_\mathrm{v}$ values (from 0 to 1 in increments of 0.1) and $c$ values (from 0 to 100 in increments of 10), which generated a single grid of size $10 \times 10$. Then we tested the learned RL policy across the entire $10 \times 10$ grid of non-policy parameters, to make predictions about model crossing decisions in all fourteen experiment scenarios for each parameter value combination. 

Building on this exhaustive parameter exploration, we determined the $\sigma_\mathrm{v}$ and $c$ that fitted the experimental data best for each participant by likelihood maximisation. We estimated the probability density function (PDF) of CIT predicted for each model by kernel density estimation, separately for each of the fourteen scenarios (\tablename~\ref{tab:scenarios}). This allowed us to calculate the model likelihood of each $\sigma_\mathrm{v}$ and $c$ for each participant, as the product of multiplying the model-predicted PDF values at the participant’s observed button press times.



    
    



\section{Results}
\label{sec:Results}

\subsection{Empirical results}
\label{subsec:Experimental results}

\begin{figure}[!t]
    \vspace{-0.2cm}
      \centering
      \includegraphics[scale=0.4]{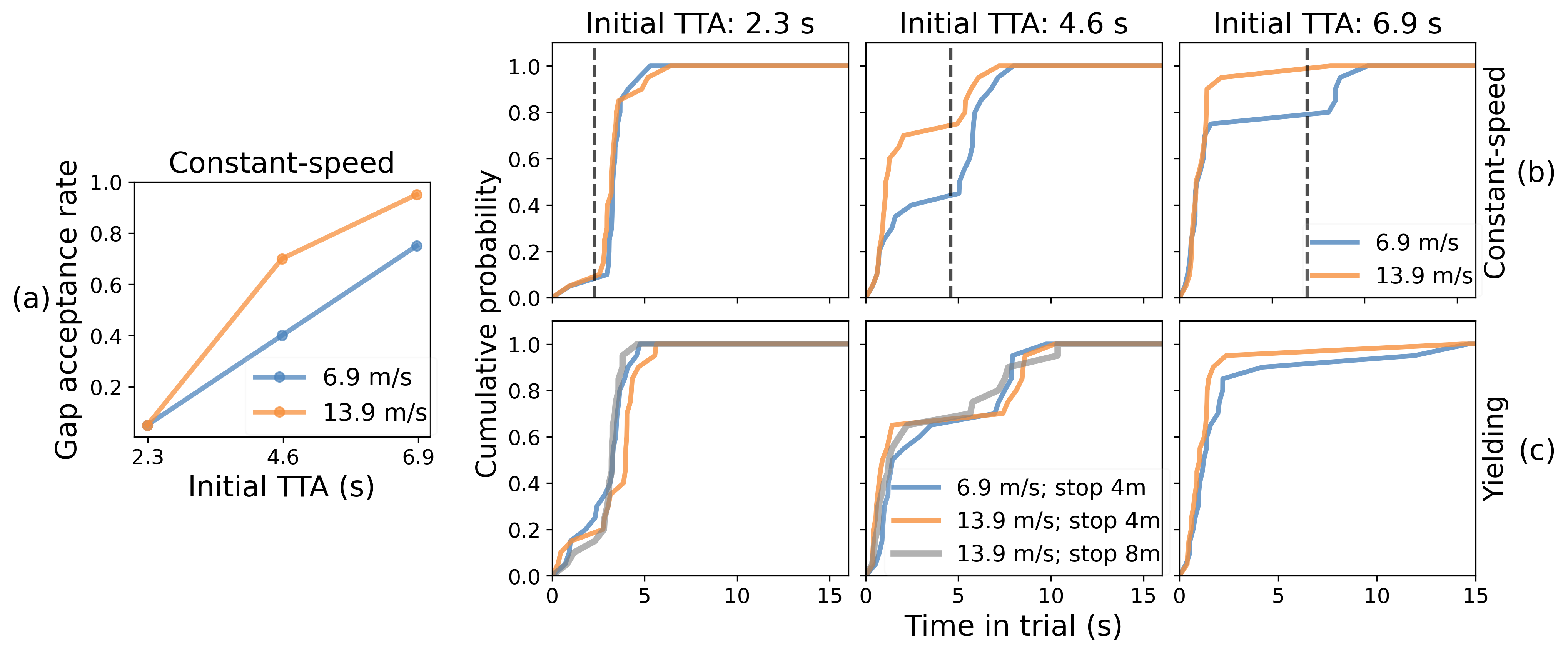}
       \vspace{-0.3cm}
      \caption{Empirical results. (a) Gap acceptance rate in constant-speed scenarios. (b) Cumulative probability of Crossing Initiation Time (CIT) in constant-speed scenarios; black dashed vertical lines indicate the times vehicles passed pedestrians. The X-axis means the time elapsed from the beginning of each trial. (c) Cumulative probability of CIT in yielding scenarios. Note: In yielding scenarios, there are only two conditions in the initial TTA of 6.9 s as shown in \tablename~\ref{tab:scenarios}, so there is no gray line in the third panel of yielding scenarios.}
      \label{fig:Exp}
      \vspace{-0.2cm}
\end{figure}

Panel (a) of~\figurename~\ref{fig:Exp} shows the gap acceptance rate observed in the scenarios where the vehicle maintained a constant speed, as a function of the speed and of the initial TTA of the vehicle. Aligning with previous research \citep{oxley2005crossing,lobjois2007age,petzoldt2014relationship}, the pedestrian gap acceptance rate was largely determined by the initial TTA (p $=$ 0.000037 and p $<$ 0.00001 for 6.9 m/s and 13.9 m/s respectively; Chi-squared test). In addition, panel (a) of~\figurename~\ref{fig:Exp} illustrates a speed-dependent gap acceptance rate, suggesting that pedestrians were more likely to accept the gap as the vehicle's speed increased, given the same initial TTA, which aligned with the finding from previous studies \citep{lobjois2007age,tian2022explaining}. However, while this speed effect was particularly evident at initial TTA of 4.6 s and 6.9 s, the associated p-values (0.057 and 0.077) are not statistically significant (p$<$0.05), possibly due to our limited sample size. 

The same speed dependency is visible also in the CIT metric \citep{tian2022explaining}, which in our case denotes the time from the start of each trial to the pedestrian's crossing decision. The cumulative probability of CIT in constant-speed scenarios is presented in panel (b) of~\figurename~\ref{fig:Exp}. We can observe a speed-dependent mean CIT (the blue cumulative probability curve is below the orange curve, indicating a higher mean CIT in the 6.9 m/s scenario), which was most obvious in the initial TTA of 4.6 s (p $=$ 0.00005; Wilcoxon Signed-Rank test). This speed effect in the constant-speed scenario is effectively the same as the one seen in the gap acceptance metric.

As for the yielding scenarios, where the car stopped to let the participant cross, it can be seen in \figurename~\ref{fig:Exp} (c) that the car speed's impact on CIT was most pronounced under a higher initial TTA condition of 6.9 s (p $<$ 0.00005; Wilcoxon Signed-Rank test). Specifically, the third graph in panel (c) of~\figurename~\ref{fig:Exp} shows the orange curve ascending more rapidly than the blue curve, which means that more pedestrians cross earlier when the initial speed of the vehicle is higher at higher TTA conditions. This speed-dependent trend is the same as in constant-speed scenarios. Conversely, for a lower initial TTA of 2.3 s, an inverse pattern emerged. More pedestrians chose to cross earlier when the approaching vehicle's initial speed was low during yielding (p $<$ 0.00015; Wilcoxon Signed-Rank test), indicating a speed-dependent yielding acceptance. This speed-dependent yielding behaviour is illustrated by the blue curve's quicker increase compared to the orange curve in the first graph of the panel (c) in~\figurename~\ref{fig:Exp}. This finding is in line with the work by \cite{tian2023deceleration}, suggesting that pedestrians tend to interpret low speed in itself as indicative of vehicle yielding.

Interestingly, a combination of these two phenomena is observed at the initial TTA of 4.6 s. In the early phase of the scenario, the CIT curves show a speed-dependent gap acceptance behaviour similar to that observed at an initial TTA of 6.9 s. However, in the later phase of the trial, a speed-dependent yielding acceptance behaviour emerges, akin to that at an initial TTA of 2.3 s. In other words, higher car speeds elicit lower CITs in the early phase of the scenario, but higher CITs in the later phase, such that across the entire scenario, the mean CIT is not significantly affected by car speed (p = 0.206; Wilcoxon Signed-Rank test).

Finally, stopping distance-dependent yield acceptance can also be seen in \figurename~\ref{fig:Exp}. More pedestrians tended to cross early when the vehicle decelerated at a higher rate, as shown by the orange line lagging behind the gray line in the yield acceptance phase of the scenario (from about 2 s in the TTA 2.3 s scenario, and from about 5 s in the TTA 4.6 s scenario), indicating that greater vehicle deceleration encourages early crossing. This effect of stopping distance on CIT was statistically significant at the initial TTA of 2.3 s (p = 0.005; Wilcoxon Signed-Rank test).

In sum, the empirical observations show indications of all of the four main phenomena we are targeting, even if some of them are present as trends rather than as statistically significant effects in this relatively small dataset.

\subsection{Model results}

\label{subsec:Model results}
After observing these key behaviours in empirical settings, we now investigate whether these phenomena are reproduced by our model. 

All model variants were found to converge; a plot of the reward curves can be found in \ref{subsec:Model convergence}.

\subsubsection{Evaluation of model outcomes against observed phenomena}
\label{subsubsec:Evaluation of model outcomes against observed phenomena}

We evaluated our models based on their ability to replicate our four targeted key pedestrian behaviours identified in empirical studies.

\begin{figure}[!t]
      \vspace{-0.2cm}
      \centering
      \includegraphics[scale=0.32]{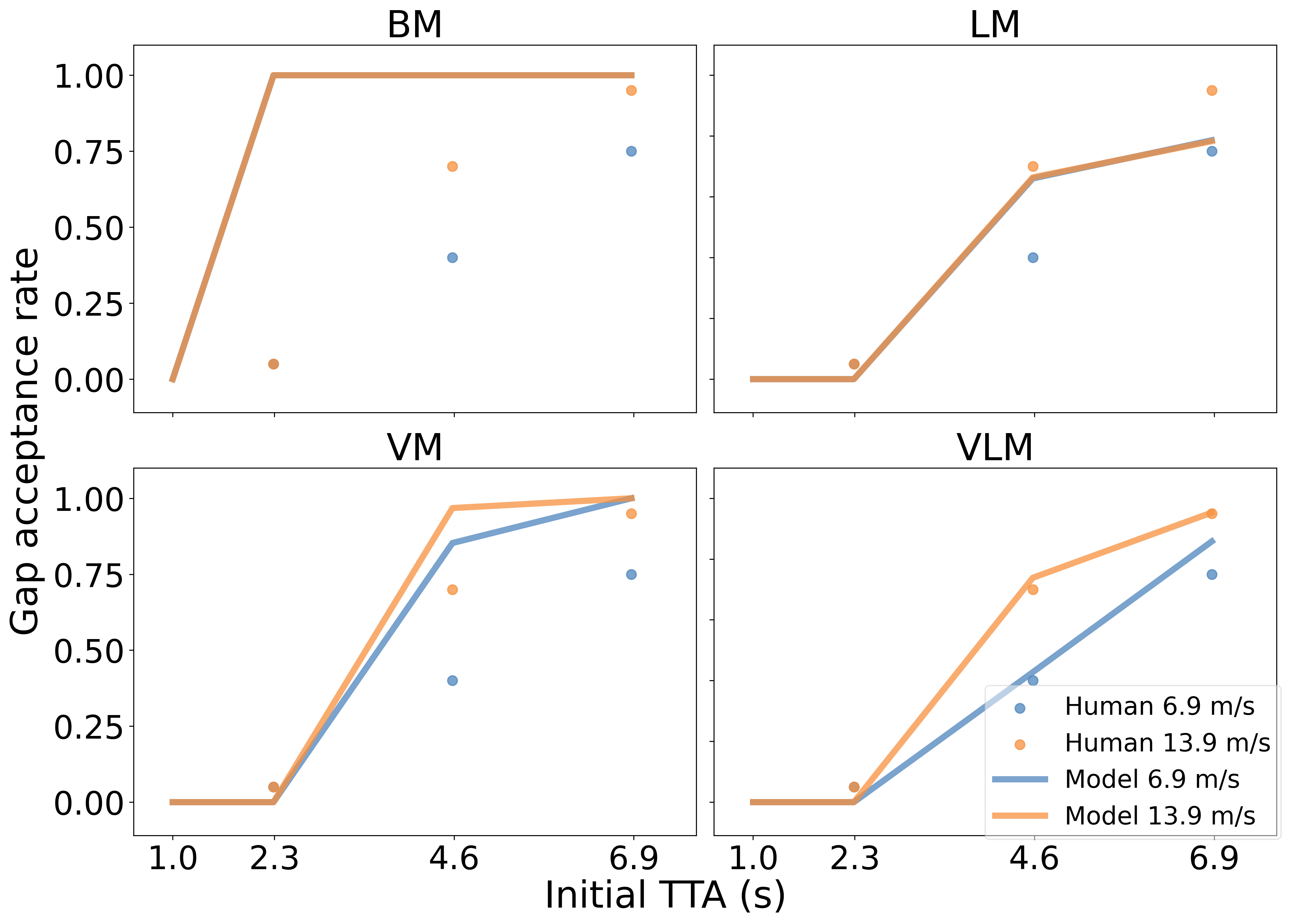}
       \vspace{-0.2cm}
      \caption{Gap acceptance rate by human participants and different models. BM: Baseline model. LM: Model with the looming aversion assumption. VM: Model with the visual limitation assumption. VLM: Model with both visual limitation and looming aversion assumptions.
      Note: the lines of the model LM and BM overlap each other.}
      \label{fig:GapPlot}
      \vspace{-0.2cm}
\end{figure}

\begin{figure}[!t]
      \vspace{-0.2cm}
      \centering
      \includegraphics[scale=0.4]{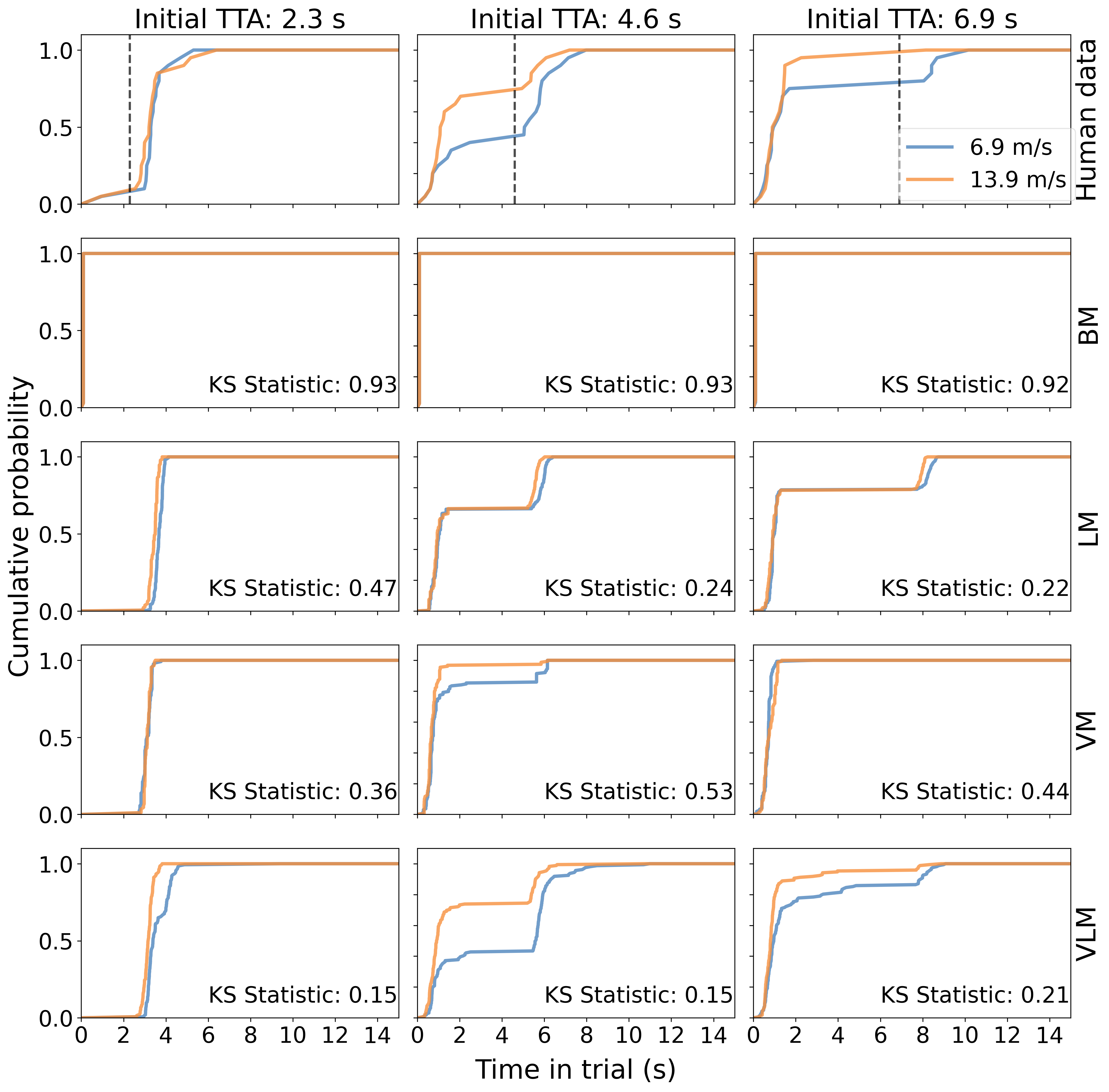}
      \vspace{-0.7cm}
      \caption{Cumulative probability for Crossing Initiation Time (CIT) in constant-speed scenarios. Black dashed vertical lines in Human data indicate the times vehicles passed pedestrians. To avoid repetition, these grey lines are not included in the model results. Vehicles passed pedestrians at the same time in the trial for each initial TTA condition. The X-axis means the time elapsed from the beginning of each trial. The KS statistic quantifies the maximum divergence between the cumulative distribution functions of human data and model results, indicating their distributional similarity. Model abbreviations are as defined in \figurename~\ref{fig:GapPlot}.}
      \label{fig:CIT_Constant}
      \vspace{-0.2cm}
\end{figure}

\begin{figure}[!t]
      \vspace{-0.2cm}
      \centering
      \includegraphics[scale=0.4]{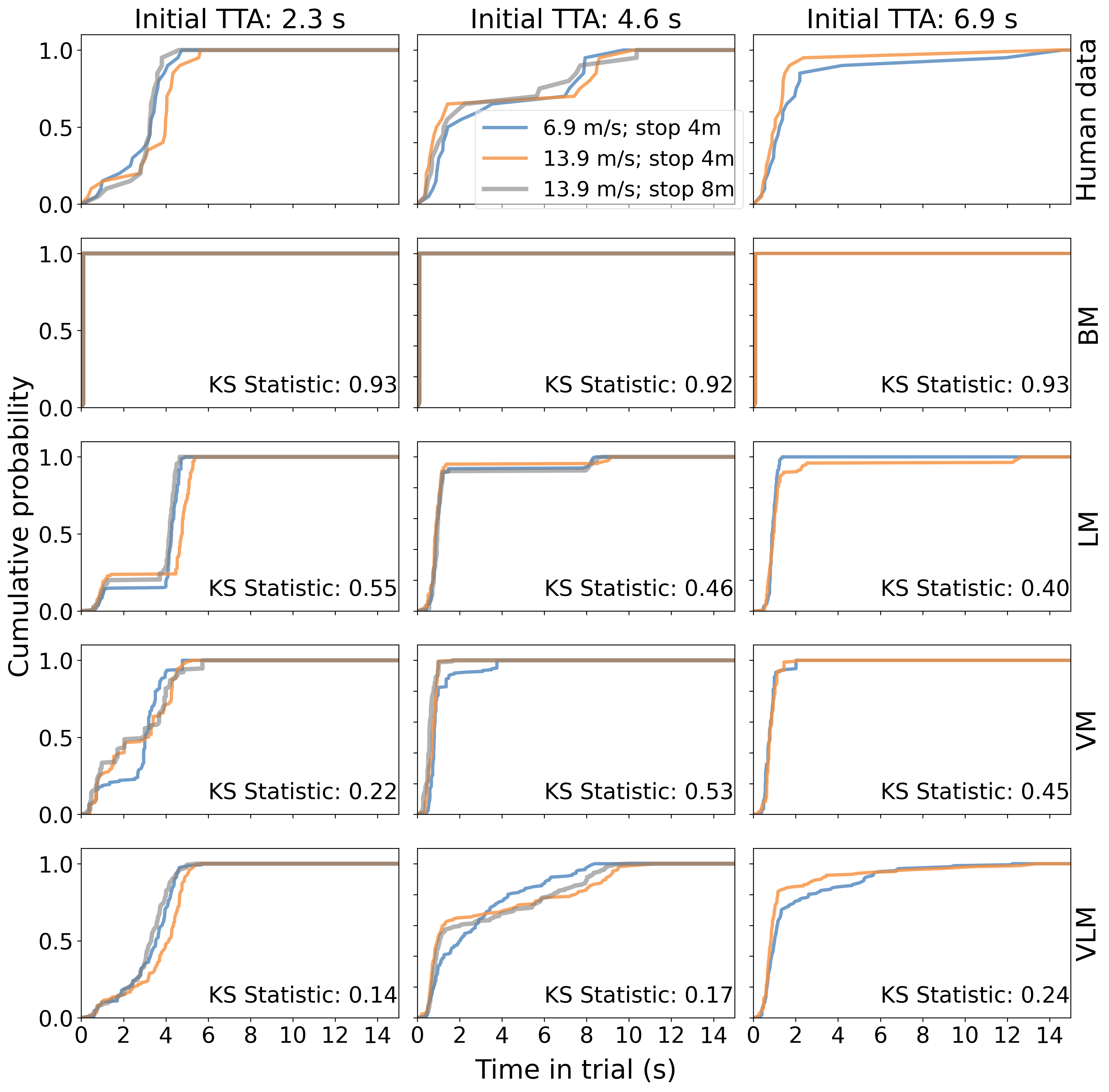}
      \vspace{-0.7cm}
      \caption{Cumulative probability for CIT in yielding scenarios. The KS statistic is explained in detail in \figurename~\ref{fig:CIT_Constant}.}
      \label{fig:CIT_Dec}
      \vspace{-0.2cm}
\end{figure}

First, we analysed the results of the baseline model (BM). The results of BM for the constant-speed scenarios are shown in the upper-left panel of \figurename~\ref{fig:GapPlot} and the second row in \figurename~\ref{fig:CIT_Constant}. As shown in \figurename~\ref{fig:GapPlot}, the BM agent decided not to cross at the initial TTA of 1 s, but consistently accepted all safe crossing opportunities available during initial TTA conditions 2.3 s, 4.6 s, and 6.9 s, where it crossed at the very start of the trial (shown in \figurename~\ref{fig:CIT_Constant}). This pattern remains the same for the yielding scenarios (\figurename~\ref{fig:CIT_Dec}). This behaviour can be attributed to the agent's perfect information about the vehicle's state, permitting it to achieve optimal behaviour for an agent whose goal is to cross the road safely in the least possible time. The BM model variant does not capture any of the four targeted phenomena. Below we will now consider these phenomena one at a time.

\paragraph{TTA-dependent gap acceptance}

Both the VM and LM models qualitatively captured the TTA-dependent gap acceptance phenomenon, as they showed increasing gap acceptance rates with rising initial TTA conditions. This pattern is particularly evident in the LM model, as can be seen in the upper-right panel of \figurename~\ref{fig:GapPlot}. The VLM model, integrating both visual limitations and looming aversion (shown in the bottom-right panel of \figurename~\ref{fig:GapPlot}) most accurately reflected the human data, demonstrating a clear sensitivity to initial TTA conditions.

\paragraph{Speed-dependent gap acceptance}
While the inclusion of just looming aversion (model LM) was enough to capture the effect of TTA on gap acceptance, this model did not show any effect of vehicle speed on gap acceptance. This can be seen in \figurename~\ref{fig:GapPlot}, where the LM lines for the two speed conditions overlap entirely. On the other hand, the inclusion of visual limitations (model VM) caused the model to exhibit speed-dependent gap acceptance behaviour, in the form of a separation of the gap acceptance curves at the initial TTA of 4.6 s. The VLM model again exhibited the best similarity with human data, showing a clearly speed-dependent gap acceptance rate. 
 
\begin{table}
    \vspace{-0.4cm}
    \caption{Quantitative assessment of model performance}
    \centering
    \begin{tabular}{ccccc}
        \toprule
         \textbf{Model type} & \textbf{Log Lik.} & \textbf{Params.} & \textbf{AIC} & \textbf{MAD(s)} \\
         \midrule
        BM  & $-1289$  & $0$ & $2578$ & $2.90$  \\
        LM  & $-594$  & $20$ & $1227$ & $0.94$  \\
        VM  & $-588$  & $20$ & $1215$ & $1.30$  \\
        VLM  & $-533$ & $40$ & $1146$ & $0.34$ \\
        VLM(E) & $-547$ & $2$ & $1098$ & $0.36$ \\
        \bottomrule
        \multicolumn{5}{l}{\footnotesize Note: Log Lik. refers to Log Likelihood, Params. refers to Free Parameters,} \\
        \multicolumn{5}{l}{\footnotesize and MAD refers to the mean absolute deviation across all scenarios.}
    \end{tabular}
    \label{tab:assessment}
    \vspace{-0.4cm}
\end{table}

\begin{figure}[b!]
      \vspace{-0.2cm}
      \centering
      \includegraphics[scale=0.38]{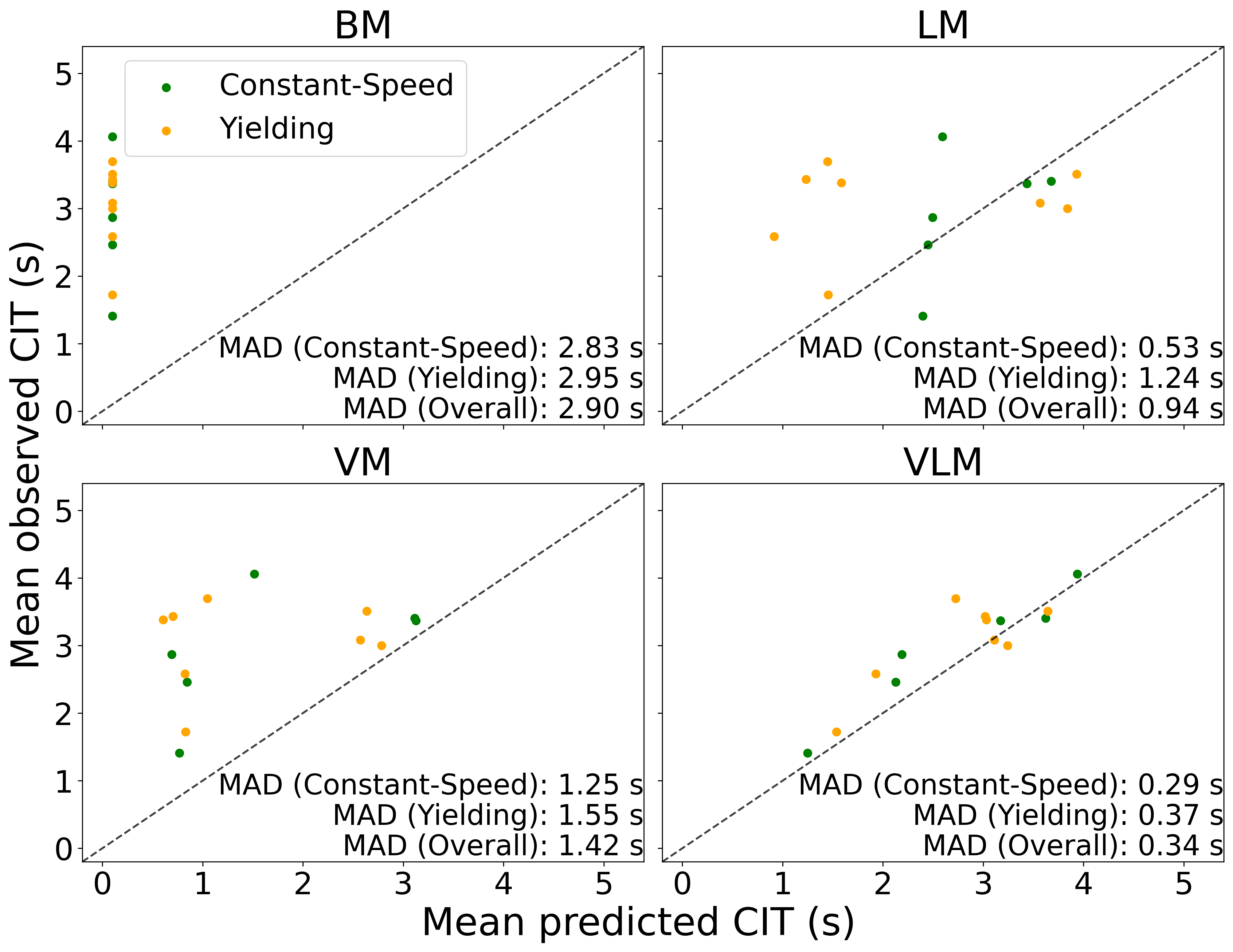}
      \vspace{-0.3cm}
      \caption{Predicted vs observed mean Crossing Initiation Time (CIT) across the different scenarios. This comparison illustrates the model's performance in estimating CIT in relation to actual measurements, with the dotted line representing an ideal prediction where observed and predicted values match perfectly.}
      \label{fig:MAD}
      \vspace{-0.2cm}
\end{figure}

\paragraph{Speed-dependent yielding acceptance}
As shown in the bottom row in~\figurename~\ref{fig:CIT_Dec}, at the initial TTA condition of 2.3 s, speed-dependent yielding acceptance emerged in the results of VLM, similar to the human data, visible as a quicker increase of the blue curve than the orange curve. In other words, the agent was more likely to cross the road earlier when the initial speed of the approaching vehicle was lower. Furthermore, and again in line with the human data, at the initial TTA condition of 6.9 s, the initial speed of the vehicle has the opposite effect on CITs (speed-dependent gap acceptance), and in the 4.6 s initial TTA condition both of the two phenomena are visible. These speed effects can to some extent be observed in the results of LM and VM at the initial TTA of 2.3 s as well, but not for the 4.6 s and 6.9 s initial TTA conditions. 

\paragraph{Stopping distance-dependent yielding acceptance}
The VLM agent replicated the short-stopping phenomenon both at the initial TTA condition of 2.3 s and 4.6 s, i.e., the agent had a greater tendency to cross earlier when the vehicle stopped at a greater distance, visible as the orange line lagging the gray line in the CIT plots. Notably, at the initial TTA of 4.6 s, this tendency appeared only in the late crossing decisions (from about 5 s). This is probably because the deceleration information has a stronger effect on the crossing decision when the vehicle is not far from the pedestrian, in line with \citep{tian2023deceleration}. The LM model also captured this pattern, especially at the initial TTA condition of 2.3 s. However, this pattern did not appear in the results of the VM model.

\begin{figure}[b!]
      \vspace{-0.2cm}
      \centering
      \includegraphics[scale=0.4]{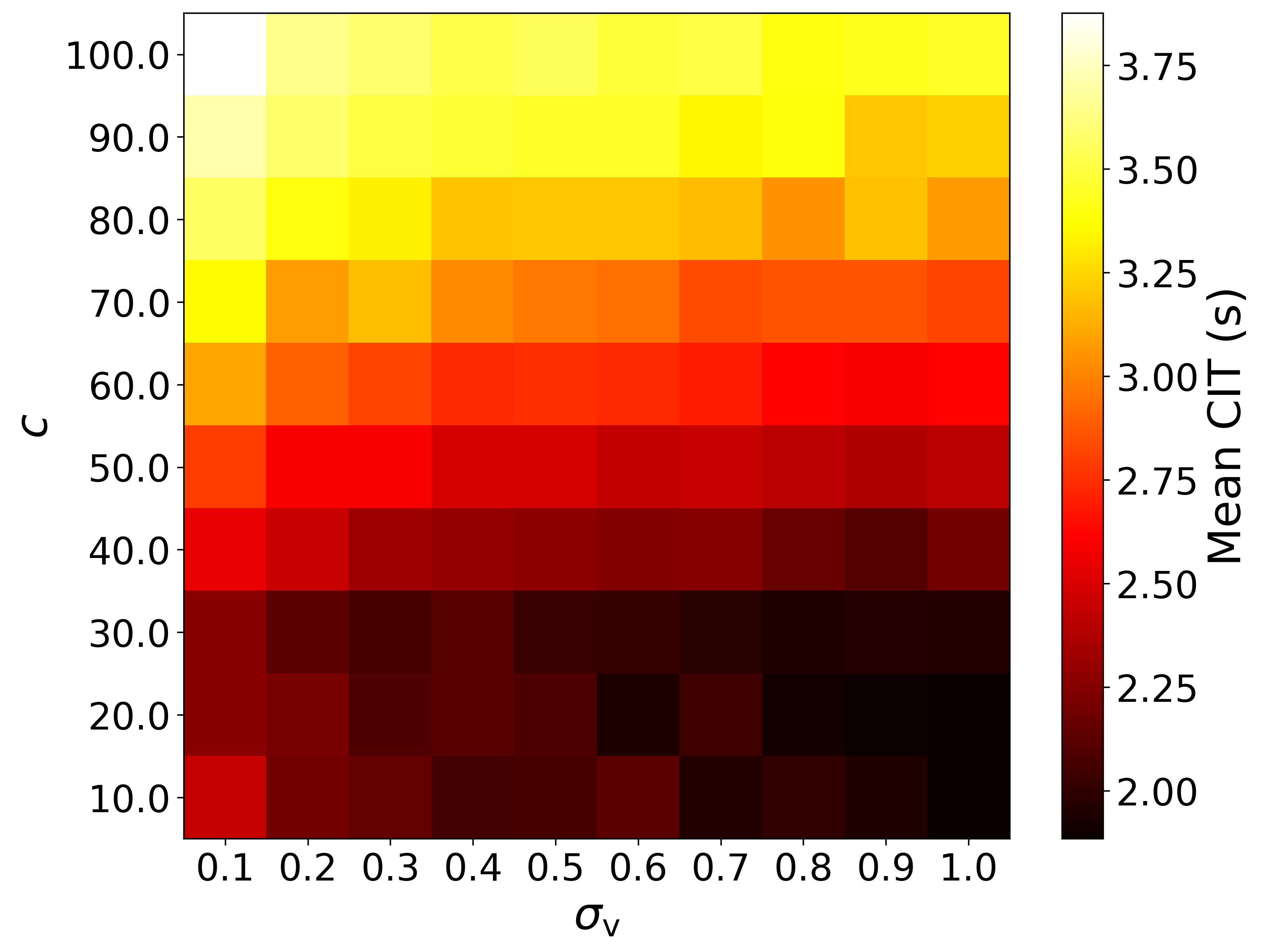}
      \vspace{-0.2cm}
      \caption{Heatmap of mean CIT for different values of the two non-policy parameters. Different colours represent the average CIT values across all different scenarios.}
      \label{fig:Heatmap}
      \vspace{-0.4cm}
\end{figure}

\subsubsection{Quantitative Assessment of Model Performance}
\label{subsubsec:Quantitative Assessment of Model Performance}

To quantitatively evaluate the performance of various models, we examined several metrics to compare observed and predicted CITs, including log-likelihood, Akaike information criterion (AIC), and mean absolute deviation (MAD). While log-likelihood values gauge the goodness of fit, AIC values provide a measure that considers both model fit and complexity \citep{farrell2008empirical}. Ideally, a good model achieves a balance between a high log-likelihood (good fit) and a low number of parameters (simplicity), resulting in a lower AIC. As shown in \tablename~\ref{tab:assessment}, the VLM model stands out among the four model variants. Not only does it have the highest log-likelihood value (-533), but also the lowest AIC score (1146). This is further corroborated by the Kolmogorov-Smirnov (KS) statistic, a nonparametric test that quantifies the maximum distance between the cumulative distribution of human data and simulation data, as presented in \figurename~\ref{fig:CIT_Constant} and \ref{fig:CIT_Dec}. A smaller KS value illustrates a closer match between the model's output and the experimental data, indicating a more accurate representation of real-world pedestrian behaviour by the model. Here, the VLM model exhibited minimal overall KS statistics, indicating that the discrepancy between the simulated and experimental CIT distributions was the smallest in the VLM model when compared to other model variants.

Regarding the MAD metric, it represents the difference between the mean predicted and observed crossing times, as illustrated in \figurename~\ref{fig:MAD}. Among all model variants, VLM outperformed all others by exhibiting the smallest deviation of just 0.34 s across all scenarios. Interestingly, all the other models tended to show faster responses than VLM, and all models showed the tendency that the model predicts more accurately in constant-speed scenarios than in yielding scenarios. 

Finally, we explored the influence of free parameters, $\sigma_v$ and the looming aversion weight $c$, on the outcomes of our most effective model, the VLM. As illustrated in \figurename~\ref{fig:Heatmap}, we computed the mean CIT of VLM across all test scenarios for every combination of parameter values. There was a noticeable increase in CIT with a larger looming aversion weight c. We observed a reverse trend with regard to the visual noise parameter, $\sigma_v$. An increment in $\sigma_v$ generally led to a decrease in CIT.

\begin{figure}[!t]
      \vspace{-0.2cm}
      \centering
      \includegraphics[scale=0.4]{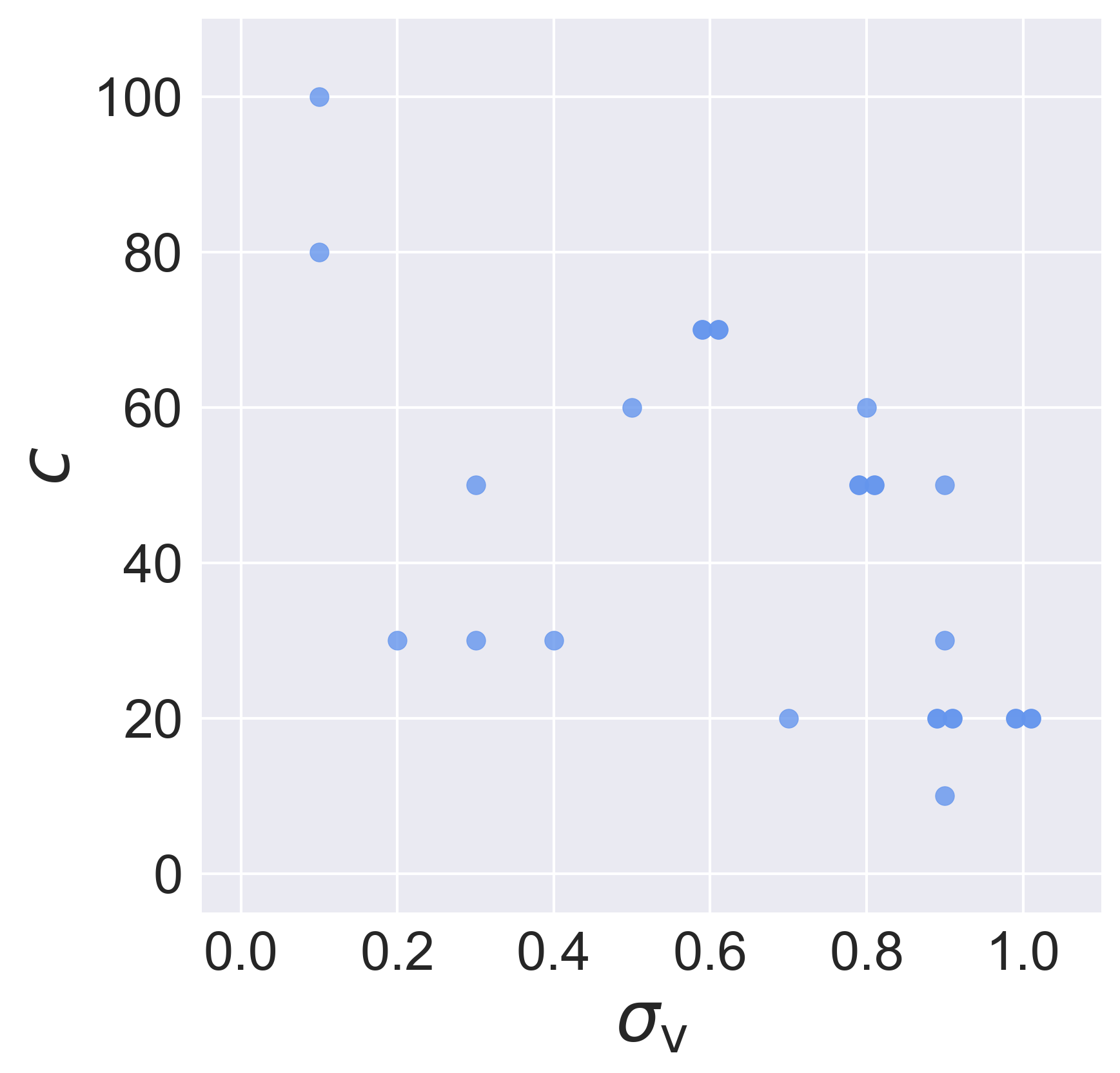}
       \vspace{-0.2cm}
      \caption{Assessment of Fitted Parameters for the model with visual limitation and looming aversion assumptions (VLM). Points represent non-policy parameter combinations, with overlapping points displayed as slightly horizontally offset to indicate multiple occurrences.}
      \label{fig:Para}
      \vspace{-0.2cm}
\end{figure}

\figurename~\ref{fig:Para} presents a scatter plot that visualises two selected non-policy parameter combinations for the VLM model, with some data points slightly offset horizontally to denote multiple occurrences. The best-fit parameters lie inside the range of the search parameters, indicating that the selected parameter range we used was sufficiently broad. 

\subsection{Additional model tests}
\label{subsec:Additional model tests}

In addition to our main analyses, we conducted tests on four additional alternative models (\ref{subsec:Additional_Model_Variants}). These include a variant that treats the motor delay as another free, non-policy parameter, a version of the looming aversion model with a different formulation of the reward function, a model with a single shared set of non-policy parameters fitted across all participants, and a variant where the RL policy was individually trained for each specific set of non-policy parameters. Among these additional model variants, only the model with a single set of non-policy parameters across the entire human data, referred to as VLM(E) in \tablename~\ref{tab:assessment} showed an advantage over the VLM variant: Due to its lower number of free model parameters it achieved a smaller AIC than VLM, despite its slightly worse goodness of fit in terms of loglikelihood and MAD.

As for the other three additional model variants, they were all outperformed by VLM. Detailed results and figures from the tests of these alternative model variants can be found in \ref{subsec:Additional_Model_Variants}.

\section{Discussion}
\label{sec:Discussion}
\subsection{Main Findings}
\label{subsec:Main Findings}
In this study, we developed a pedestrian crossing decision model. This model is based on the computational rationality framework and uses RL to derive boundedly optimal behaviour policies under assumptions about the human constraints and preferences. We see three main findings:

First, the VLM model variant, which incorporates both the visual limitation and looming aversion assumptions, captured all four of our targeted identified behavioural phenomena. To our knowledge, no previous models have captured all four of these phenomena.

Second, the necessity of including both visual limitation and looming aversion assumptions in the VLM model variant is another key insight from our research. The model variants with only one of these assumptions, VM and LM, each only reproduced some of the phenomena. This finding highlights the complexity of pedestrian decision-making processes, in which our results suggest that both visual limitations and looming aversion play integral roles.

Third, another important finding is the effectiveness of our parameter conditioning method. Applying this method to the VLM model allowed us to fit the model to the empirical data and allowed us to demonstrate good quantitative fits with just two free model parameters for the entire dataset (or two parameters per participant, if fitting per individual).

In the following subsections, we will discuss the impact of our mechanistic assumptions about human constraints and preferences on crossing decisions, and explore the broader implications and limitations of our study.

\subsection{Impact of mechanistic assumptions on crossing decisions}
\label{subsec:Impact of Cognitive Constraints on Crossing Decisions}
In this subsection, we will first discuss how assumptions about human constraint and preference affect the generation of the four targeted phenomena:

\paragraph{TTA-dependent gap acceptance}
While both VM and LM models demonstrate increased gap acceptance with rising initial TTA, this effect is more evident in the LM model compared to the VM model. This finding can be taken to suggest that the looming aversion plays an important role in the assessment of safe crossing opportunities. In the LM model, a greater TTA reduces \(\frac{1}{\tau}\), resulting in a lower looming aversion reward for crossing at higher TTA conditions. This phenomenon aligns with the model's inherent logic, where a higher TTA is associated with a lower perceived threat, encouraging the agent to cross.

Regarding VM, the qualitative pattern of TTA-dependent gap acceptance for this model is caused by the perceptual noise. Given that the agent cannot observe the state of the vehicle accurately, there is a higher risk of collision during crossing, especially in the lower TTA conditions. To avoid the penalty from the collision, the agent is more conservative and less likely to cross at the low initial TTA condition (2.3 s). However, at higher initial TTAs, the impact of the perceptual noise is less detrimental to making safe crossing decisions, leading to higher gap acceptance rates.

\paragraph{Speed-dependent gap acceptance} Our findings indicate that models incorporating the visual limitation assumption are capable of generating speed-dependent gap acceptance behaviour. It is not entirely trivial to understand why speed-dependent gap acceptance arises from the visual limitation. Our best interpretation is that it is related to the more dispersed distribution of estimated TTA at lower vehicle speeds for the given time gap, as shown in  \figurename~\ref{fig:TTA_Dist}. Therefore, from a reward maximisation perspective, the agent decides to make more cautious decisions when interacting with a lower-speed vehicle.

This finding is interesting in relation to past explanations of speed-dependent gap acceptance as being caused by biases in human TTA estimation, due to humans making  inappropriate use of distance information \citep{petzoldt2014relationship} or visual looming information \citep{tian2022explaining} when judging crossing safety. Our findings instead suggest that speed-dependent gap acceptance is an optimal behaviour, but that this is an optimality which is bounded by the particular noise characteristics of the human visual system. Differently put, given the nature of human visual perception, speed-dependent gap acceptance is an entirely rational behaviour. It is of course still possible that human pedestrians make use of visual looming information when making these decisions, as suggested by \citet{tian2022explaining}, but if so this seems to be part of a boundedly optimal strategy, rather than a biased (i.e., suboptimal) heuristic.

\paragraph{Speed-dependent yielding acceptance} Our results show that both the model with the noisy perception assumption and the model with the looming aversion assumption can independently generate the speed-dependent yielding acceptance behaviour. This can be understood as follows: Under the noisy perception assumption, perceptual noise decreases as the distance between the vehicle and the agent reduces. When a vehicle approaches at a slower speed, for a given TTA it is closer to the pedestrian, resulting in less perceptual noise. Consequently, the agent can more accurately estimate the vehicle's position, leading to an earlier decision to cross. For the model with looming aversion, the slower vehicle speed near the agent reduces the \(\frac{1}{\tau}\) value. This lower value increases the reward for crossing, thus encouraging the agent to cross earlier during the vehicle's yielding.

\paragraph{Stopping distance-dependent yielding acceptance} This behaviour is captured effectively by the looming aversion assumption (but not by the noisy perception assumption). This can be understood by considering that when a vehicle decelerates from the same initial distance and speed, stopping at a greater deceleration further away from the pedestrian results in a smaller \(\frac{1}{\tau}\), thereby reducing the penalty to the agent for crossing in front of the vehicle. Therefore, the agent has a greater tendency to cross when the vehicle stops further away.

\paragraph{Exploration of non-policy parameters} As illustrated in~\figurename~\ref{fig:Heatmap}, there is a noticeable increase in average CIT with a larger looming aversion weight $c$. This is reasonable since a higher value of $c$ amplifies the perceived threat, thus leading the agent to make more cautious crossing decisions.

Conversely, higher noise levels $\sigma_\mathrm{v}$ in the visual system are associated with slight reductions in mean CIT. This outcome is somewhat counterintuitive, especially when contrasted with the smaller CIT exhibited by BM, which operates without visual noise. This disparity suggests a complex interaction between the looming aversion and visual noise parameters in VLM’s decision-making process. A possible explanation could be that lower levels of visual noise allow the agent to assess looming aversion more accurately, and thus more strategically delay crossing the road to avoid the looming aversion penalty. Conversely, when the level of visual noise is high, the agent's risk assessment of potential collisions becomes less reliable, which may lead to a tendency to cross the road more quickly to avoid the collision and time penalties, despite the penalty for looming aversion, and thus maintain a lower overall penalty from the perspective of reward maximisation.

\subsection{Implications and limitations}
\label{subsec:Implications and limitations}

Different from previous studies in modelling work \citep{pekkanen2021variable,tian2022explaining}, we do not pre-define how the agent should make decisions. Instead, our model allows the agent to learn boundedly optimal decision-making, given its environment and constraints. By continually encountering varied scenarios in the dynamic environment, the agent learns its policy through an iterative process of trial and error. This dynamic learning approach bears several advantages. For one, it imparts a level of adaptability to our model, enabling it to be attuned to different possible real-world conditions. Furthermore, the inherent adaptability of RL positions this kind of model favourably for application in road user behaviour modelling more broadly, which is a key motivation behind our work. Given the great success of modern deep RL in learning behaviour policy for highly challenging tasks, it can be argued that the approach we have taken here is extensible to modelling of road user behaviour and interaction also in very general, high-complexity traffic scenarios. This is an interesting direction for future work, which will require both adoption of more advanced RL methods than we have used here, and importantly also further research to establish more complete mechanistic models of the involved human perceptual, cognitive, and motor constraints. 

While our learning-based approach is not just a computational novelty but an embodiment of how humans naturally adapt and refine their judgments based on experiences, it is crucial to note that the learning process of the RL policy in our model is not intended to mimic the exact learning process in humans. While the model converges towards a policy that could be similar to boundedly optimal human behaviour, the path it takes to reach this convergence is different from human adaptation. The value of this RL approach lies in its ability to explore complex scenarios and generate behaviour that aligns with the principles of computational rationality, thereby providing a partly mechanistic and more explainable alternative to purely data-driven ML models.

Furthermore, our findings contribute to the accuracy of predicting human decision-making in pedestrian crossings. We reveal that two parameters – the looming aversion weight and the noise of agents' visual systems – have a salient influence on the modelled CIT. In this regard, using real-world data from a large representative population to ascertain these parameters, and subsequently integrating them into RL processes, may be important steps towards enhanced model accuracy.

The method of training the RL model conditioned on non-policy parameters is established in the realm of computational rationality within the HCI domain \citep{keurulainen2023amortised}. Yet, applying this method to pedestrian modelling represents a novel and promising direction. This approach is particularly beneficial in navigating extensive parameter spaces, suggesting a future of more tailored pedestrian behaviour models. Again, applying this model to real-world pedestrian data presents a promising research direction. For example, we could infer the non-policy parameters unique to different individuals, such as their risk perception or walking speed, before they interact with vehicles. This could lead to more precise prediction of pedestrian behaviours in various traffic scenarios. 

In the field of AV simulation, the potential applications of our model are promising. For instance, it could improve the realism and effectiveness of AV testing simulations \citep{Markkula2022Models}. Furthermore, by incorporating a detailed understanding of pedestrian behaviour, real-time AV algorithms can be fine-tuned to anticipate and respond to pedestrian actions more accurately. For example, by combining our model's predictions with real-time data from vehicle sensors, AV systems could achieve a higher level of situational awareness, enabling them to make safer decisions in complex urban environments \citep{camara2020pedestrian}. Similar uses of these models in for example infrastructure design can also benefit pedestrian traffic safety more broadly, also in conventional human-driven traffic \citep{kraidi2020pedestrian,zhu2022effect}.

There is clear and ample scope for further improvements to our model. For example, there is more variability in the human CITs than in the model CITs. This could be due to both between- and within-individual variability of decision-influencing factors. In addition, our reward function is simple and so far, we have fine-tuned only parts of it (the looming aversion). This was sufficient for our purposes, to show qualitative patterns of human road-crossing, and contrary to our expectations it even allowed us to achieve relatively good quantitative fits, but there is clear room for further work. For example, following a similar approach to what we did here with $\sigma_\mathrm{v}$, we could tune also for example the time-loss penalty in the reward function. This adjustment would allow the model to more accurately reflect the different ways pedestrians value the trade-off between time and safety.

Another aspect to consider for future improvement of our model is its current focus exclusively on the pedestrian's reaction to vehicles, without accounting for the driver's reaction to the pedestrian, which is mentioned in Section~\ref{sec:Methods}. Real-world scenarios often involve a dynamic interplay between these parties, influencing each other’s decisions. To address this gap, future studies could employ multiagent RL to model the interactions between pedestrians and drivers more realistically \citep{hu1998multiagent}.

Finally, it is interesting to note that we face the challenge of fully understanding the complexity of our model's behaviour. This issue arises both in relation to the speed-dependent gap acceptance (we have shown that this behaviour is boundedly optimal, but we have not been able to fully explain why) and the effect of visual noise levels on CITs (zero noise causes rapid decisions, but low levels of noise cause slower decisions than high levels of noise). While our computational rationality model is more interpretable than fully data-driven ML models, it does not entirely eliminate the challenge of explicating the 'why' behind its decision-making processes. On a high level, the behaviour of our model, governed by the principle of computational rationality and informed by RL, can always be explained quite simply as boundedly optimal, but formulating a more detailed explanation of \emph{why} a certain behaviour is boundedly optimal may not always be straightforward.

\section{Conclusions}
\label{sec:Conclusions}

In conclusion, this study offers a contribution to the field of modelling human pedestrian crossing decisions, utilising a computational rationality framework, and using RL to identify boundedly optimal behaviour policy. We demonstrate that our model, under our chosen human-like constraints, can emulate human-like behaviour in road-crossing scenarios with both non-yielding and yielding vehicles. Interestingly, our findings suggest that the previously reported speed-dependent gap acceptance behaviour in pedestrian decision-making can be understood as a rational adaptation to visual perception limitations. Our best model not only reproduces this behaviour but also qualitatively and quantitatively captures three other key phenomena: TTA-dependent gap acceptance, speed-dependent yielding acceptance, and stopping distance-dependent yielding acceptance. In addition, we identified two parameters - the weight of looming aversion and the noise in agents' visual systems - that have an influence on CIT, which can provide a better understanding of human decision-making in real-world pedestrian crossing scenarios. While we recognise the need for further investigate and develop our model, the insight from this study provides a promising foundation for improved pedestrian models. Not least, the inherent adaptability of RL suggests the potential for future extensions to more complex traffic scenarios, a challenge that traditional mechanistic models may find difficult to address. The potential applications of our model are notable. In the field of AVs, our results can contribute to more realistic road users in virtual testing environments and guide vehicular decision-making algorithms to better anticipate human behaviour, ensuring safer human-AV co-existence. 

\section*{CRediT authorship contribution statement}

\noindent \textbf{Yueyang Wang}: Conceptualization, Methodology, Formal analysis, Software, Investigation, Validation, Writing -- original draft, Writing -- review \& editing. \textbf{Aravinda Ramakrishnan Srinivasan}: Conceptualization, Methodology, Investigation, Formal analysis, Validation, Writing -- review \& editing, Supervision. \textbf{Jussi P.P. Jokinen}: Conceptualization, Methodology, Writing -- review \& editing. \textbf{Antti Oulasvirta}: Conceptualization, Methodology, Writing -- review \& editing. \textbf{Gustav Markkula}: Conceptualization, Methodology, Investigation, Formal analysis, Validation, Writing -- review \& editing, Supervision.

\section*{Acknowledgments}
This work was supported by the UK Engineering and Physical Sciences Research Council (grant EP/S005056/1); the Academy of Finland (grant 330347); and the University of Leeds International Academic Mobility Scheme.

\section*{Declarations of interest: none}


\newpage
\appendix
\section{Additional Material}
\label{sec:appendix}

\subsection{Model convergence}
\label{subsec:Model convergence}

\begin{figure}[h]
      \vspace{-0.2cm}
      \centering
      \includegraphics[scale=0.4]{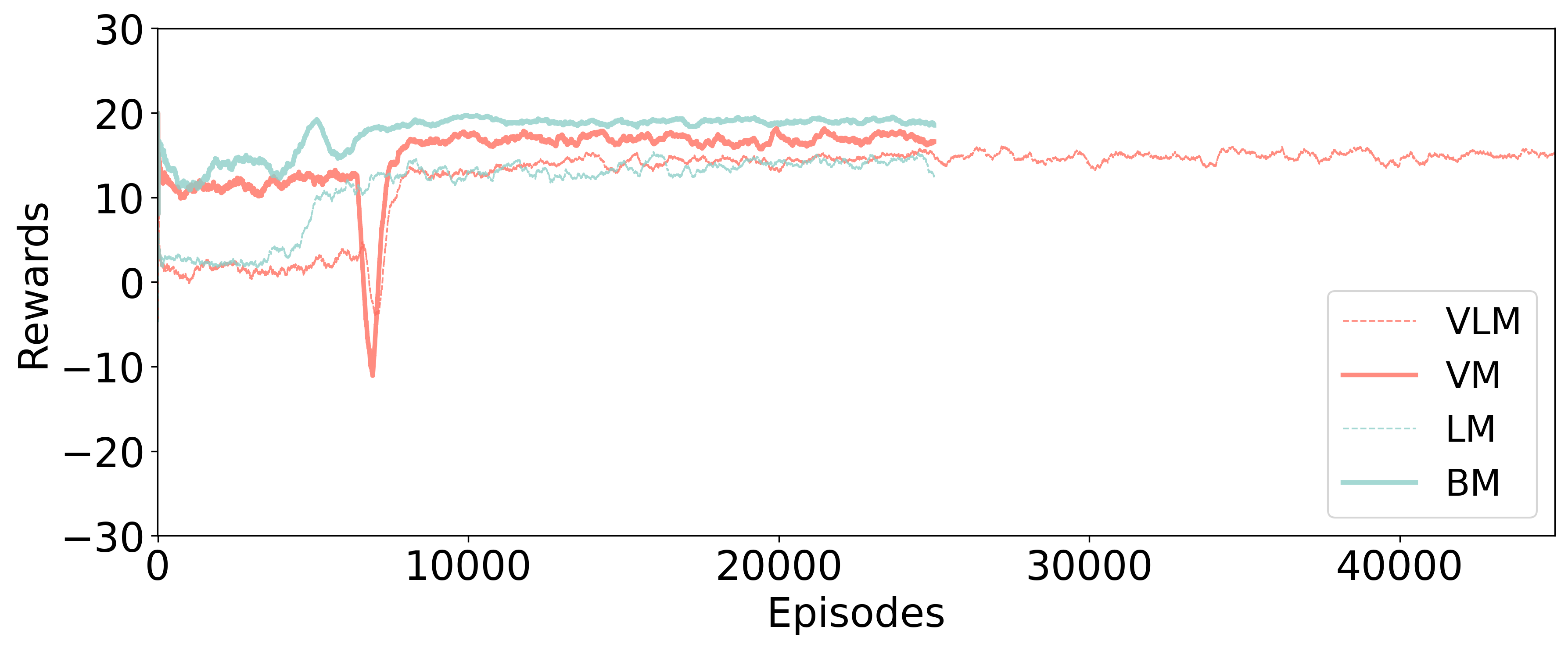}
       \vspace{-0.2cm}
      \caption{Convergence plot for different models.}
      \label{fig:Reward}
      \vspace{-0.2cm}
\end{figure}
\figurename~\ref{fig:Reward} illustrates the average rewards of four main model variants recorded for each of the 500 episodes, spanning the entirety of the training phase. From \figurename~\ref{fig:Reward}, it can be noted that the reward for the models including the looming aversion assumption (LM and VLM) achieve lower rewards overall, due to the change in the reward structure introduced by this assumption. Additionally, adding visual limitations to the baseline model (going from BM to VM) led to a decline in the average rewards. This is due to the VM agent's inability to know the precise position of the approaching vehicle. As a result, the agent tends to make more cautious decisions, leading to increased time penalties. However, there was no noticeable difference in the achievable rewards for LM and VLM. A possible reason could be that for LM, the looming aversion penalty is directly proportional to the inverse $\tau$, while in VLM, the looming aversion penalty corresponds to the estimated inverse $\tau$. From the reward maximisation perspective of view, both model variants aim to minimise this looming penalty within their respective environmental contexts, which potentially explains the similar overall rewards.

\subsection{Additional Model Variants}
\label{subsec:Additional_Model_Variants}

Beyond the core model variants outlined in Section~\ref{subsec:Main Model variants}, we explored four additional variations to deepen our understanding of the pedestrian decision-making process:

\begin{itemize}
    \item \textbf{VLDM:} The model variant that considers motor delay as an additional human-like constraint, reflecting the time lag between decision-making and action execution. This variant expands upon the VLM model variant, incorporating motor delay into the model.
    \item \textbf{VNM:} A variation of the looming aversion model that tests an alternative representation of risk aversion. Instead of an inverse $/tau$, we evaluated a model where aversion is formulated against near-collision situations, providing an alternative perspective on risk assessment in pedestrian behaviour.
    \item \textbf{VLM (E):} A version of the VLM model that employs a single set of non-policy parameters across the entire human dataset, instead of fitting them to each participants. This approach aimed to assess the efficacy of a more generalised model application.
    \item \textbf{VLM (S):} This variation of the VLM model involves training individual networks for each unique set of parameters.
\end{itemize}

\begin{table}[b]
    \caption{Quantitative assessment of additional model performance}
    \centering
    \begin{tabular}{ccccc}
        \toprule
         \textbf{Model type} & \textbf{Log Lik.} & \textbf{Params.} & \textbf{AIC} & \textbf{MAD(s)} \\
         \midrule
        VLDM  & $-536$  & $60$ & $1192$ & $0.35$  \\
        VNM  & $-564$  & $40$ & $1208$ & $0.99$  \\
        VLM(E) & $-547$ & $2$ & $1098$ & $0.36$ \\
        VLM(S)  & $-572$  & $40$ & $1225$ & $0.39$  \\
        \bottomrule
    \end{tabular}
    \label{tab:Appendix_assessment}
\end{table}

\begin{figure}[!t]
      \centering
      \includegraphics[scale=0.4]{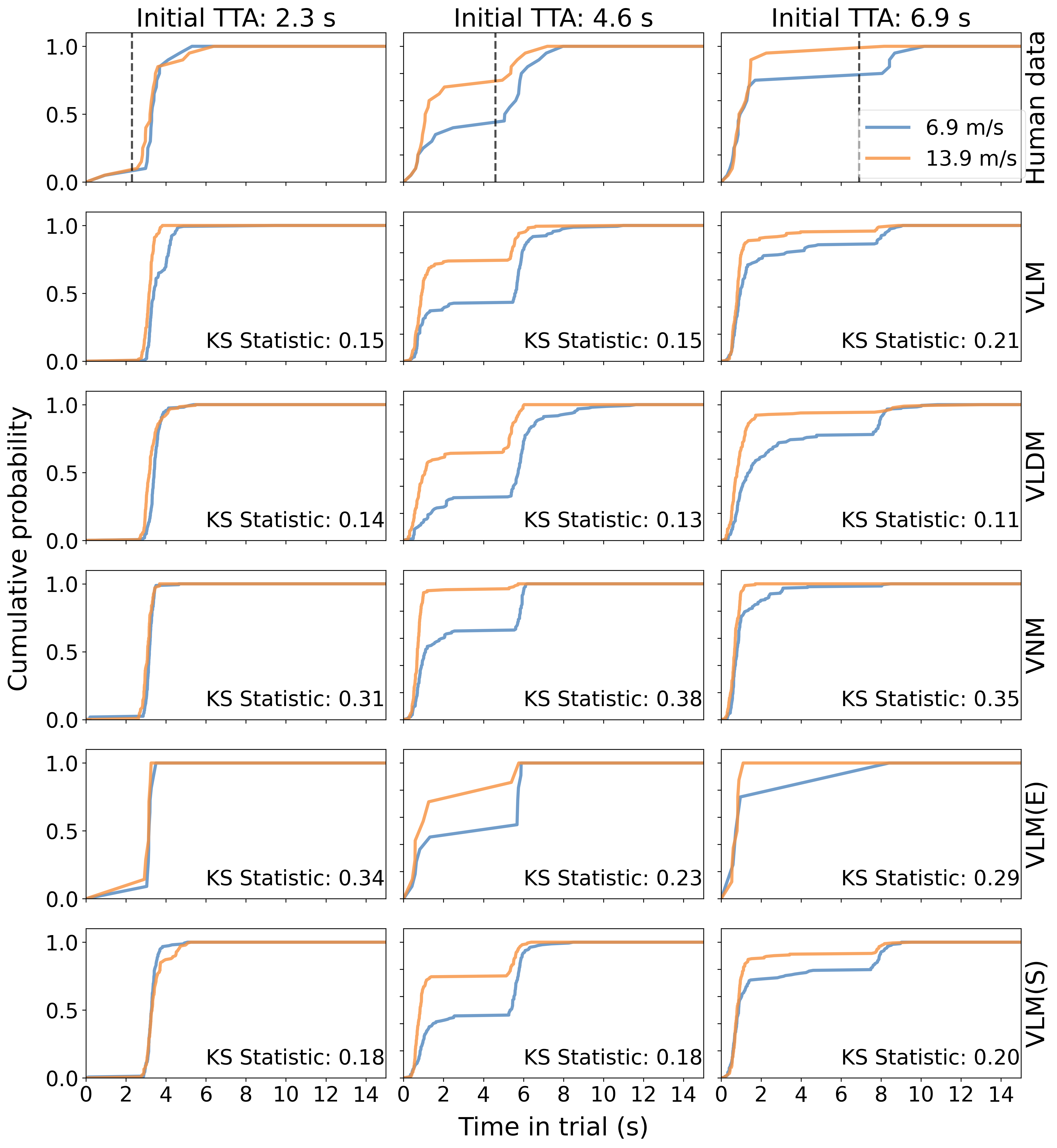}
      \vspace{-0.7cm}
      \caption{Cumulative probability for Crossing Initiation Time (CIT) in constant speed scenarios for human data and additional models. Black dashed vertical lines in Human data indicate the times vehicles passed pedestrians. To avoid repetition, these black lines are not included in the model results. Vehicles passed pedestrians at the same time in the trial for each initial TTA condition. The VLM model is included in this figure for comparison.}
      \label{fig:CIT_ConstantAlternative}
      \vspace{-0.2cm}
\end{figure}

\begin{figure}[!t]
      \vspace{-0.2cm}
      \centering
      \includegraphics[scale=0.4]{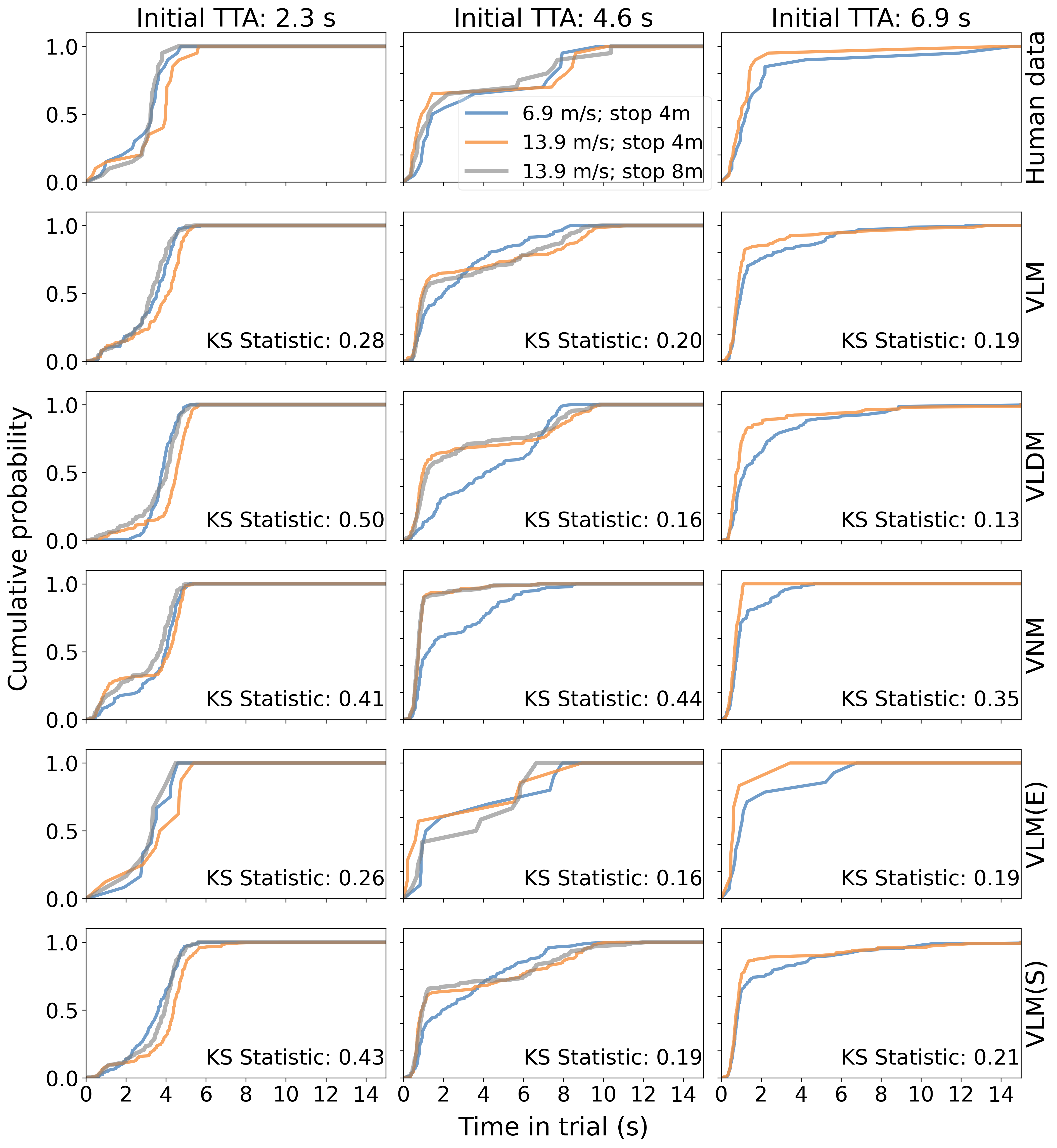}
      \vspace{-0.7cm}
      \caption{Cumulative probability for CIT in yielding scenarios for human data and additional models. The VLM model is included in this figure for comparison.}
      \label{fig:CIT_DecAlternative}
      \vspace{-0.2cm}
\end{figure}

These models were designed to assess the influence of additional factors and assumptions on pedestrian crossing behaviour. While these models provided valuable insights, they did not show an overall improvement in performance compared to our primary model, VLM. ~\figurename~\ref{fig:CIT_ConstantAlternative} and \ref{fig:CIT_DecAlternative} and \tablename~\ref{tab:Appendix_assessment} show the results obtained for these models.

The VLDM model integrates motor delay ($m$) as an additional non-policy parameter, alongside visual limitations ($\sigma_\mathrm{v}$) and looming aversion ($c$). VLDM showed similar performance to the VLM model in terms of log-likelihood and MAD. However, it obtained a higher AIC score due to the increased model complexity.

\begin{figure}[!t]
      \vspace{-0.3cm}
      \centering
      \includegraphics[scale=0.35]{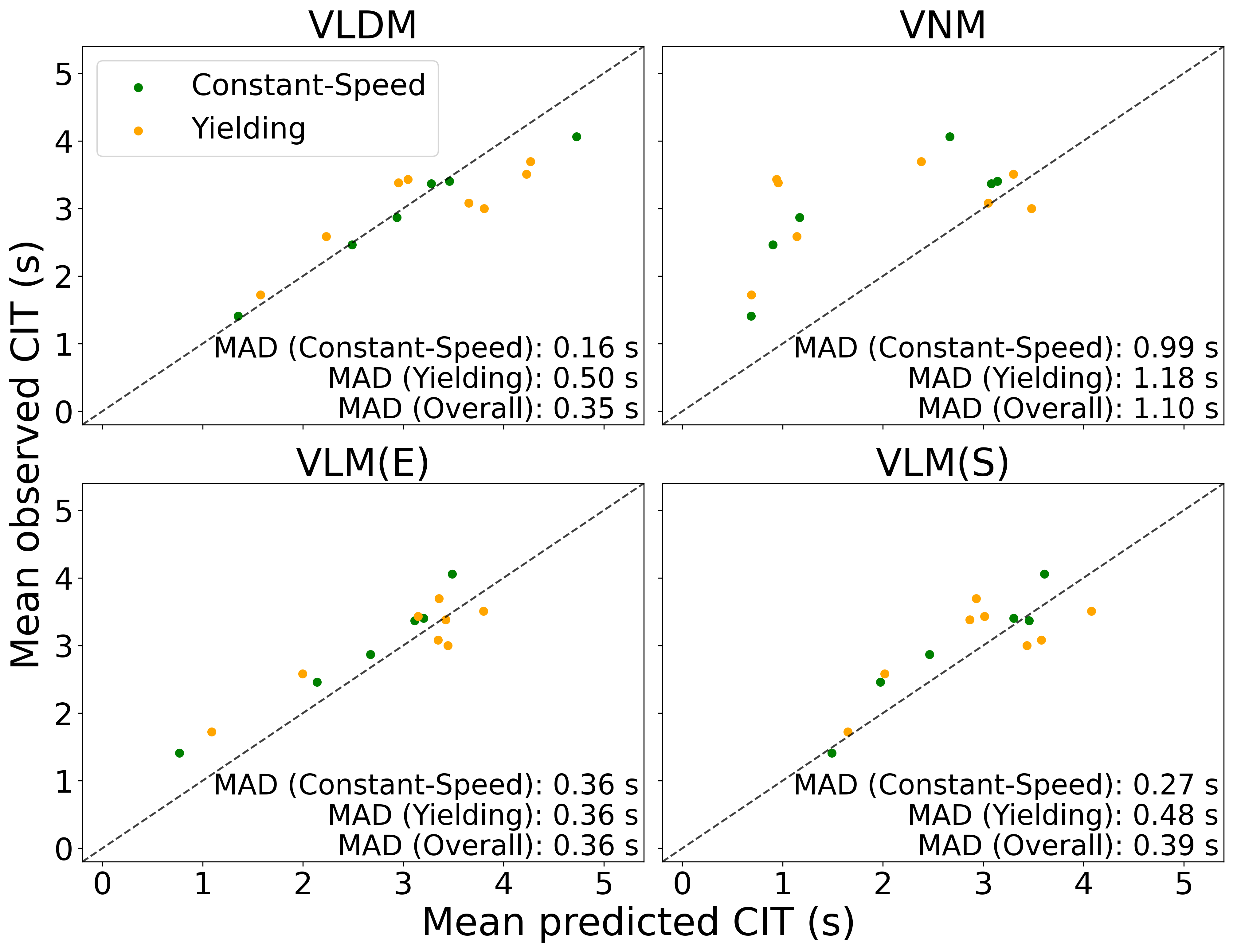}
      \vspace{-0.2cm}
      \caption{Predicted vs observed mean CIT across the different scenarios.}
      \label{fig:MAD_Alt}
\end{figure}

The VNM model explores the concept of pedestrian aversion to near-collision situations. This model variant provides an alternative way of modelling pedestrian aversion to being in safety-critical situations, as an alternative to the formulation based on looming in the LM/VLM models. This model assumes pedestrians inherently aim to avoid situations where a vehicle invades their personal space, leading to an additional penalty in the model when this occurs. While VNM exhibited more human-like CITs than the VM model, it still lagged behind the comprehensive performance of the VLM model regarding AIC. This disparity in model performance could imply that the integration of temporal-spatial clues (e.g., looming) may play a more important role in pedestrian decision-making than spatial factors alone (e.g., proximity to a near-collision zone).

\begin{figure}[!t]
      \centering
      \includegraphics[scale=0.55]{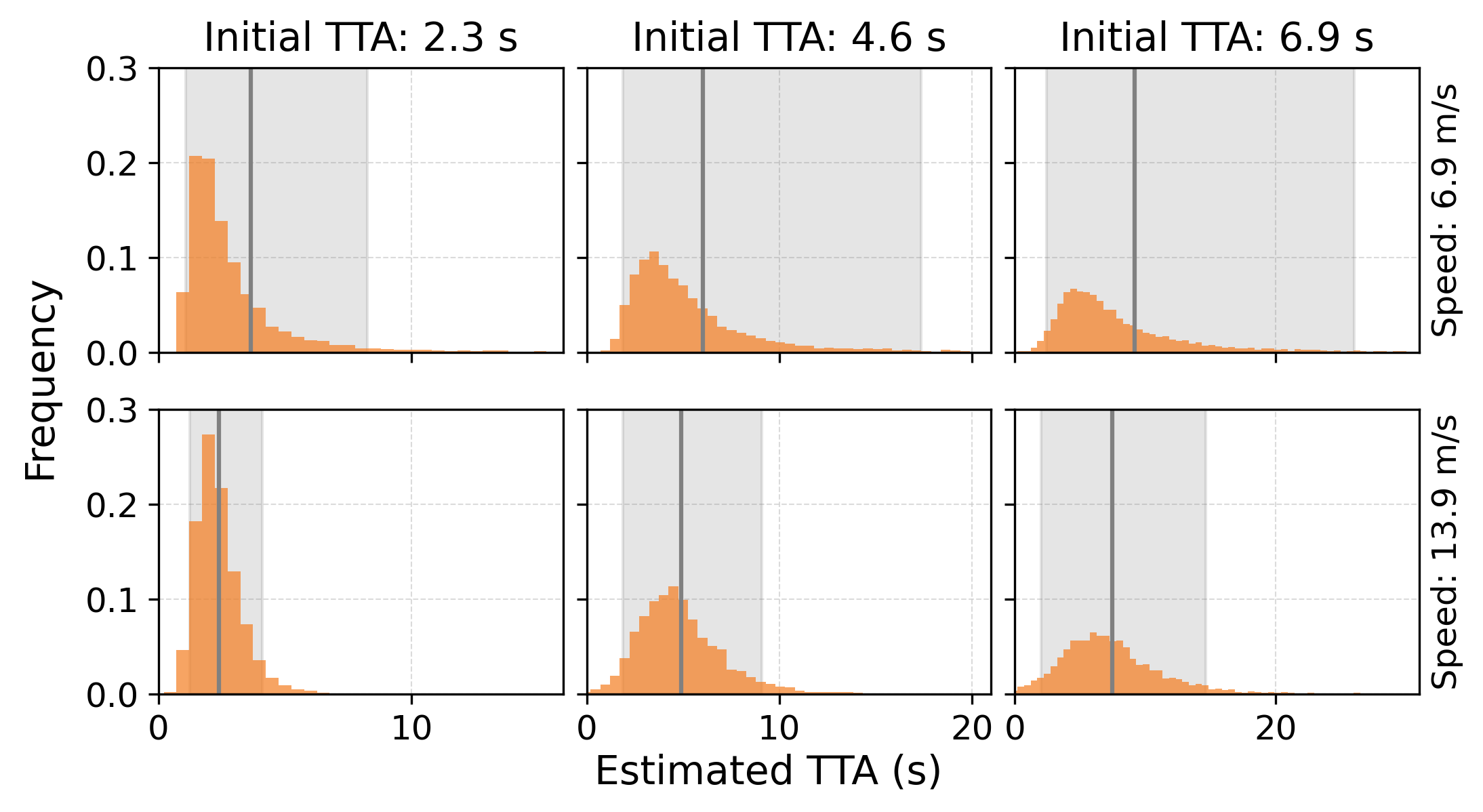}
      \vspace{-0.2cm}
      \caption{Distributions of initial estimated TTA from the output of the visual perception model, across the different scenarios in the experiment. The grey vertical line is the mean value, and the shading area shows the 5th and 95th percentiles of the estimated TTA.}
      \label{fig:TTA_Dist}
      \vspace{-0.2cm}
\end{figure}

The VLM (E) model, which uses a single set of non-policy parameters for the entire dataset, performed better in terms of AIC value compared to VLM, reflecting a simpler model structure. However, it did not match the log-likelihood performance of VLM. This finding suggests that individualising non-policy parameters may be beneficial for accurately capturing individual pedestrian behaviour.

In the VLM (S) model, instead of conditioning a single RL policy on the values of the non-policy parameters, we trained separate RL policy networks for each of the 10x10=100 considered values of the non-policy parameters. Despite the increased computational effort, VLM (S) did not achieve better performance metrics than the original VLM model, as shown in \tablename~\ref{tab:Appendix_assessment}. This result highlights the efficiency of our chosen parameter-conditioning in the VLM model.

Regarding VLDM, with the motor delay, $m$, as an additional input parameter alongside $\sigma_\mathrm{v}$ and $c$, from \tablename~\ref{tab:Appendix_assessment}, we can find that this model matched the VLM's performance in terms of both log-likelihood and MAD. However, its additional parameters result in a higher AIC compared to VLM.

\clearpage
\bibliographystyle{elsarticle-harv} 
\bibliography{cas-refs}





\end{document}